\documentclass{article}

\PassOptionsToPackage{numbers,sort}{natbib}
\usepackage[eandd, preprint]{neurips_2026}
\usepackage[utf8]{inputenc}
\usepackage[T1]{fontenc}
\usepackage{hyperref}
\usepackage{url}
\usepackage{booktabs}
\usepackage{amsfonts}
\usepackage{nicefrac}
\usepackage{microtype}
\usepackage{xcolor}
\usepackage{graphicx}
\usepackage{amsmath}
\usepackage{tabularx}
\usepackage{subcaption}
\usepackage{algorithm}
\usepackage{algpseudocode}
\usepackage{adjustbox}
\usepackage{longtable}
\usepackage{caption}
\usepackage{multirow}
\usepackage{makecell}
\usepackage{threeparttable}
\usepackage{wrapfig}
\usepackage[table]{xcolor}
\usepackage[most]{tcolorbox}
\usepackage{float}
\usepackage{verbatim}
\usepackage{xcolor}
\usepackage{array}
\usepackage{tabularx}
\newcolumntype{Y}{>{\centering\arraybackslash}X}

\title{MolDeTox: Evaluating Language Model’s Stepwise Fragment Editing for Molecular Detoxification}

\author{
 \textbf{Jueon Park\textsuperscript{1}\thanks{Equal contribution.}}, \
 \textbf{Wonjune Jang\textsuperscript{2}\footnotemark[1]}, \
 \textbf{Jiwoo Lee\textsuperscript{1}}, \
 \textbf{Yein Park\textsuperscript{1,3}}, \
 \textbf{Jaewoo Kang\textsuperscript{1,3}\thanks{Corresponding author.}}
\\
 \textsuperscript{1}Korea University \
 \textsuperscript{2}Myongji University \\
 \textsuperscript{3}AIGEN Sciences
\\
\{jueon\_park, kangj\}@korea.ac.kr \\ dnjswnswkd03@mju.ac.kr
}

\begin{document}
\maketitle

\begin{abstract}
Large Language Models (LLMs) and Vision Language Models (VLMs) have recently shown promising capabilities in various scientific domain. In particular, these advances have opened new opportunities in drug discovery, where the ability to understand and modify molecular structures is critical for optimizing drug properties such as efficacy and toxicity. However, existing models and benchmarks often overlook toxicity-related challenges, focusing primarily on general property optimization without adequately addressing safety concerns.  In addition, even existing toxicity repair benchmarks suffer from limited data diversity, low structural validity of generated molecules, and heavy reliance on proxy models for toxicity assessment. To address these limitations, we propose \textbf{MolDeTox}, a novel benchmark for molecular detoxification, designed to enable fine-grained and reliable evaluation of toxicity-aware molecular optimization across stepwise tasks. We evaluate a wide range of general-purpose LLMs and VLMs under diverse settings, and demonstrate that understanding and generating molecules at the fragment-level improves structural validity and enhances the quality of generated molecules. Moreover, through detailed task-level performance analysis, MolDeTox provides an interpretable benchmark that enables a deeper understanding of the detoxification process. Our dataset is available at : \url{https://huggingface.co/datasets/MolDeTox/MolDeTox}
\end{abstract}


\section{Introduction}

Recent advances in large language models (LLMs) have significantly expanded their capabilities beyond natural language processing, enabling applications in scientific domains such as chemistry~\cite{narayanan2025training, zhao2025chemdfm} and biology~\cite{fallahpour2025bioreason, fallahpour2026bioreason, istrate2025rbio1}. 
They have also demonstrated an increasing ability to understand molecular structures and infer physicochemical properties directly from structural representations, making them promising tools for drug discovery. 
The integration of multimodal architectures, such as vision-language models (VLMs) has further accelerated this progress~\cite{li2025chemvlm, adak2025molvision}.

These advances have extended to molecular optimization, one of the most challenging tasks in drug discovery. Molecular optimization involves modifying molecules to achieve multiple objectives, such as enhancing biological activity, improving selectivity, and minimizing toxicity~\cite{yu2025collaborative}.
LLMs have also demonstrated the ability to improve molecular properties, such as drug-likeness(QED) and solubility, through prompt-based approaches or instruction tuning~\cite{nguyen2024lico, ye2025drugassist}.
In addition, benchmark datasets have been introduced to evaluate how effectively LLMs perform molecular optimization tasks~\cite{cai2025mollangbench, li2025beyond}.

However, molecular optimization specifically for toxicity mitigation remains under-explored.
Existing studies typically focus on only a small number of toxicity endpoints, such as hERG-related cardiotoxicity~\cite{ye2025drugassist}. 
Nevertheless, there have been efforts that focus exclusively on toxicity, such as ToxiMol~\cite{lin2025breaking}, which evaluates the molecule repair capabilities of multimodal large language models (MLLMs) across multiple toxicity endpoints. 
Despite its contributions, this benchmark has several limitations. It is constructed using only toxic molecules from existing datasets, resulting in limited diversity and a relatively small number of evaluation samples. More critically, the molecules generated by the evaluated MLLMs exhibit a high rate of structural invalidity, averaging approximately 50\%. Even for valid molecules, toxicity is assessed using a single proxy model, raising concerns about the reliability and robustness of the evaluation.

To address the limitations of existing benchmarks, we propose \textbf{MolDeTox}, a new benchmark for molecular detoxification. 
We first construct a novel dataset named \textit{ToxicityCliff}, where molecular pairs exhibit high structural similarity but different toxicity outcomes within the same endpoint. 
This dataset is derived from public toxicity resources, comprising approximately 52K molecular pairs across 49 toxicity endpoints, thereby reflecting diverse and realistic molecular optimization scenarios.
It captures cases where detoxification can be achieved through minimal structural modifications while preserving physicochemical properties. We represent molecules in SAFE~\cite{noutahi2024gotta}, a fragment-level representation that explicitly encodes substructure information. It facilitates the identification of toxic and non-toxic fragments and enables LLMs to perform molecular modifications at a finer granularity.

Building on this dataset, we formulate the task as a series of question–answering (QA) problems and decompose molecular detoxification into three sub-tasks: (1) identifying the toxicity-relevant fragments, (2) replacing them with non-toxic alternatives, and (3) generating the final detoxified molecule. This design is inspired by the workflow of human experts, who iteratively analyze and refine molecular structures~\cite{kalgutkar2019designing}.
For the final molecule generation step, instead of directly generating SMILES, we adopt a fragment-level generation strategy by first generating SAFE representations and then decoding them into SMILES. This approach significantly improves the structural validity of generated molecules compared to conventional methods.

Figure~\ref{fig:molDetox_overview} provides a snapshot of model performance across the MolDeTox benchmark. It shows how a diverse set of LLMs and VLMs perform on the three tasks under different settings. This figure highlights the overall difficulty of the benchmark and reveals performance variations across model types and task formulations. At the same time, our benchmark is inherently interpretable, as it decomposes detoxification into explicit intermediate steps, allowing us to analyze not only whether generation succeeds but also how errors arise across different stages.

Our contributions can be summarized as follows:
\begin{figure*}[t]
    \centering
    \includegraphics[width=\textwidth]{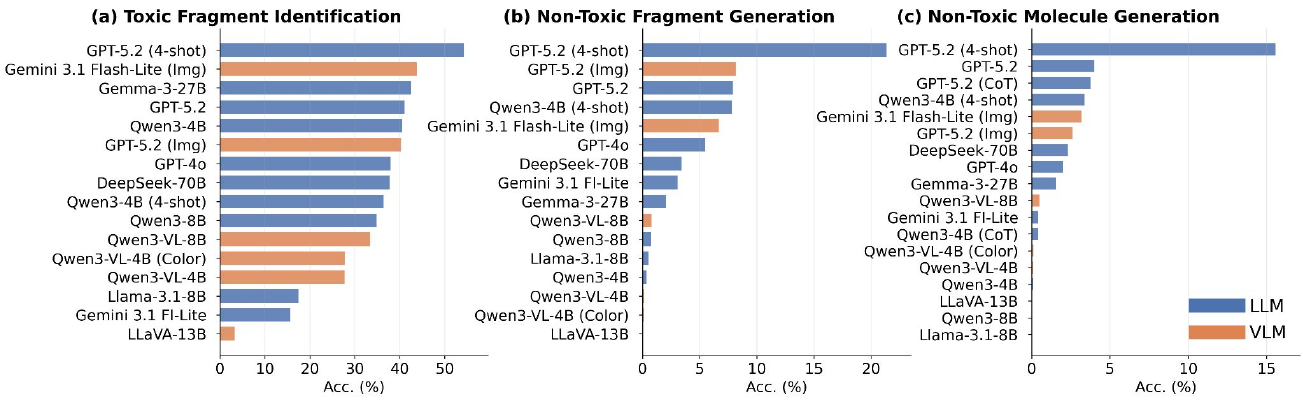}
    \caption{\emph{\textbf{MolDeTox Overview:}} Performance comparison of LLMs and VLMs across three detoxification tasks, evaluated using accuracy (\%)}
    \label{fig:molDetox_overview}
    \vspace{-15pt}
\end{figure*}

\begin{figure*}[t]
    \centering
    \includegraphics[width=\textwidth]{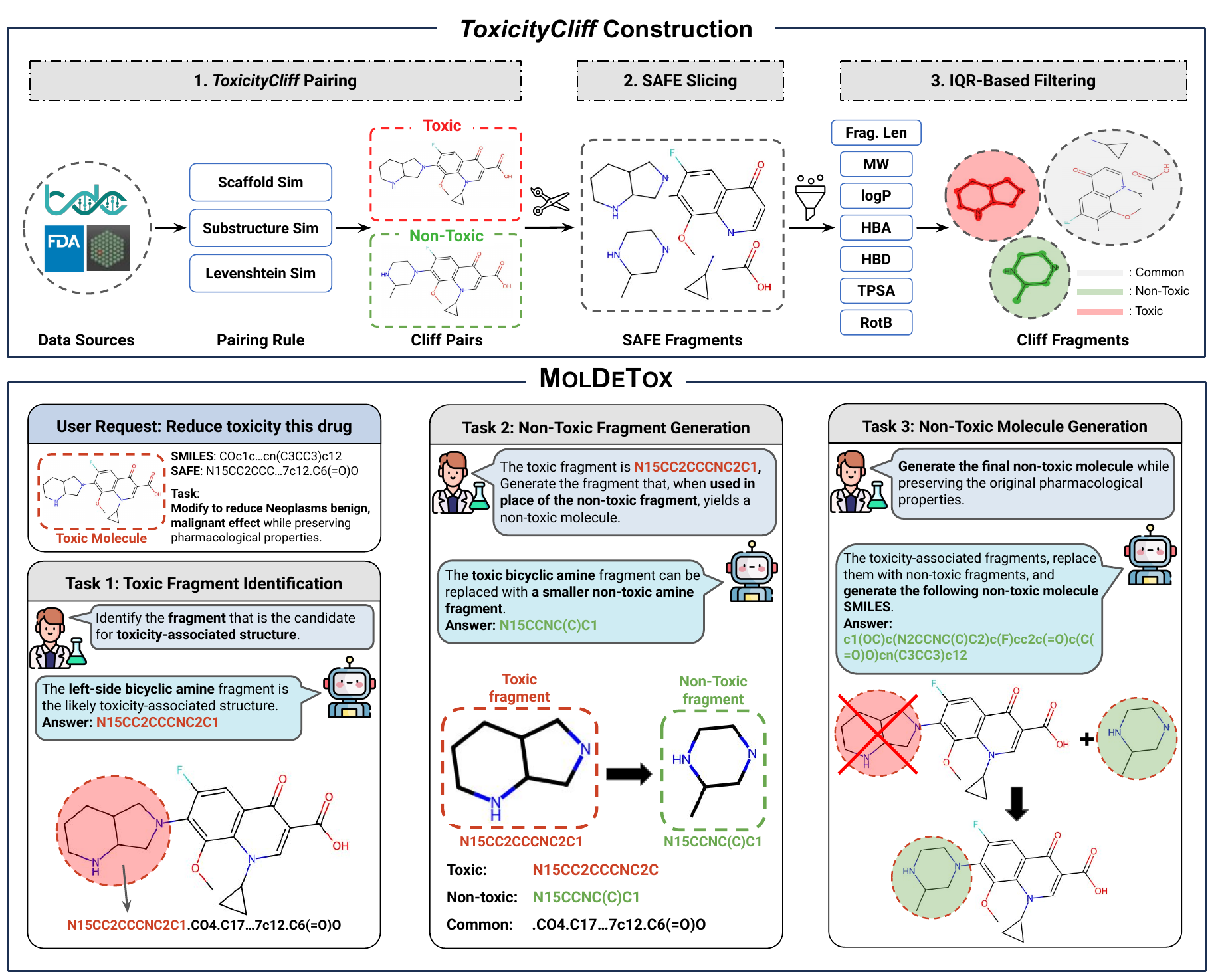}
    \caption{Pipeline of MolDeTox. The upper panel illustrates the \emph{ToxicityCliff} construction pipeline. The lower panel shows how each ToxicityCliff pair is converted into three step-wise QA tasks.}
    \label{fig:molDetox_motivation}
\end{figure*}

\begin{itemize}
    \item We construct a novel dataset, \textit{ToxicityCliff}, consisting of structurally similar molecular pairs with distinct toxicity outcomes, enabling fine-grained evaluation of molecular detoxification.
    \item We propose \textbf{MolDeTox}, a QA-based benchmark that decomposes molecular detoxification into three interpretable sub-tasks, enabling systematic evaluation of the extent to which LLMs and VLMs can mimic human expert reasoning processes.
    \item We demonstrate that using a fragment-level SAFE representation for molecular understanding and generation in both LLMs and VLMs improves structural validity and overall performance in molecular detoxification tasks.
\end{itemize}
\section{Related Work}

\paragraph{LLMs for molecular optimization}
Recent studies have explored the use of large language models (LLMs) for molecular optimization. DrugAssist \cite{ye2025drugassist} constructs a large-scale instruction dataset for molecule optimization and fine-tunes an LLM via instruction tuning. The model demonstrates multi-property optimization capabilities, including QED, solubility, and hERG inhibition. LICO\cite{nguyen2024lico} also extends LLMs for molecular optimization by performing in-context black-box optimization via learned embeddings and surrogate modeling.
More recent approaches, such as MT-M{\scriptsize OL}\cite{kim2025mt}, formulate molecular optimization as a multi-agent system combining LLMs with domain-specific tools, demonstrating strong performance via collaborative and tool-guided reasoning.

Recently, benchmark datasets have been introduced to evaluate how effectively LLMs can perform molecular optimization tasks. For example, MolLangBench\cite{cai2025mollangbench} focuses on molecular recognition, editing, and generation in language-prompted settings, reflecting real-world chemists’ workflows. In addition, ChemCoTBench \cite{li2025beyond} evaluates LLMs across two chemistry applications, including molecular optimization, where tasks are defined at both the physicochemical and target levels. However, despite extensive progress in molecular optimization models and benchmarks, toxicity mitigation remains insufficiently addressed, even though it plays a crucial role in drug discovery.

\paragraph{LLMs for molecular toxicity}
As LLMs improve their ability to reason over molecular structures, their capability to predict drug toxicity from structural representations such as SMILES has also advanced. Recent studies \cite{yang2025large, chen2025application} employ prompt-based approaches to predict specific toxicity endpoints, including cardiotoxicity and drug-induced osteotoxicity, directly from SMILES inputs. 
Furthermore, CoTox~\cite{park2025cotox} introduces a framework that integrates both chemical and biological information within a chain-of-thought reasoning paradigm, enabling the prediction of multiple organ-level toxicities.

Benchmarks have also been introduced to evaluate how well LLMs capture molecular toxicity knowledge. ToxReason~\cite{park2026toxreason} presents a benchmark for assessing LLMs’ mechanistic toxicity reasoning by requiring models to predict three organ-level toxicities (liver, heart, and kidney) based on adverse outcome pathways(AOPs).
Moreover, ToxiMol~\cite{lin2025breaking} introduces a VQA-style benchmark to evaluate whether multimodal large language models (MLLMs) can repair toxic molecules, covering 11 primary toxicity tasks and 660 representative toxic molecules. Its evaluation determines repair success based on a combination of molecular validity, QED~\cite{bickerton2012quantifying}, Synthetic Accessibility, Lipinski’s Rule of Five evaluation, Similarity, and a binary safety score predicted by TxGemma-Predict \cite{wang2025txgemma}.
In contrast, our \textbf{MolDeTox} focuses on structurally similar molecular pairs with opposite toxicity labels collected across diverse toxicity dataset. This design enables stepwise evaluation of whether LLMs and VLMs can identify and edit specific toxic fragments toward non-toxic alternatives at a finer granularity.

\section{MolDeTox}
\label{sec:methodology}

MolDeTox is designed to evaluate whether language models can follow natural language instructions for toxicity-aware molecular reasoning. Rather than treating detoxification as unconstrained molecule generation, our benchmark is built around a \emph{minimal-edit reasoning} perspective: given a toxic molecule, a model should identify toxicity-relevant fragments, propose localized non-toxic edits, and generate a non-toxic analog while preserving as much of the original molecule as possible. This formulation motivates our dataset construction pipeline, which aims to curate toxic/non-toxic molecule pairs that are globally similar yet locally different in chemically meaningful ways.

\subsection{\emph{ToxicityCliff} Construction}
MolDeTox is constructed from 10 binary-labeled drug toxicity datasets spanning 49 toxicity endpoints. We collect datasets from multiple sources, including the U.S. Food and Drug Administration (FDA), Therapeutic Data Commons (TDC), and SIDER.
From FDA resources, we include DILIst \cite{thakkar2020drug}, DICTrank \cite{qu2023dictrank}, and DIRIL \cite{connor2024generation}. From TDC, we utilize datasets from the toxicity category, including hERG blockers \cite{wang2016admet}, hERG Central \cite{du2011hergcentral}, hERG Karim \cite{karim2021cardiotox}, Ames Mutagenicity \cite{xu2012silico}, Skin Reaction \cite{alves2015predicting}, Tox21 \cite{huang2016tox21challenge}, and ClinTox \cite{gayvert2016data}. In addition, we incorporate cytochrome P450 (CYP) inhibition datasets \cite{veith2009comprehensive} from the metabolism category to capture toxicity-related metabolic effects \cite{kondvza2025targeted}. Finally, we include SIDER \cite{kuhn2016sider}, which provides drug side-effect information.

\begin{wrapfigure}{h}{0.56\textwidth}
    \centering
    \vspace{-5pt}
    \includegraphics[width=0.56\textwidth]{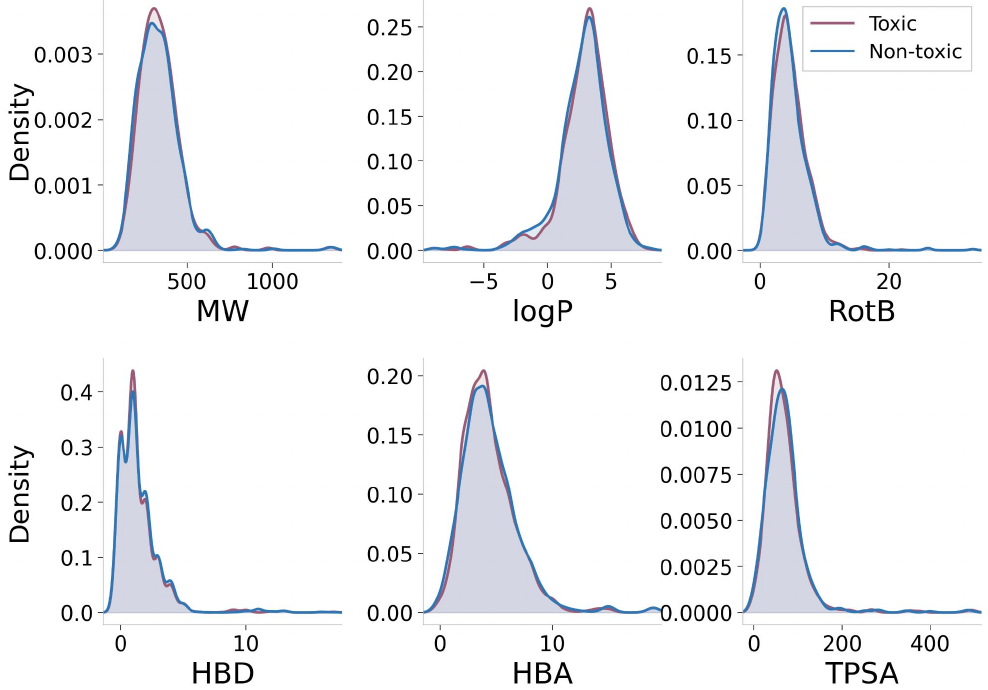}
    \caption{Property distributions of toxic and non-toxic molecules in testset.}
    \label{fig:test_property_distribution}
    \vspace{-15pt}
\end{wrapfigure}

To preserve the physicochemical properties of molecules while removing toxic effects, we leverage the concept of activity cliff to the toxicity domain. Activity cliffs, where structurally similar molecules exhibit large differences in biological properties, have been widely studied~\cite{van2022exposing, kim2025graphcliff}, and are often explained through matched molecular pair analysis (MMPA), which links small, localized structural changes to significant variations in molecular activity~\cite{kalgutkar2019designing, yang2023matched}. Specifically, for each endpoint, we identify pairs of molecules that are structurally highly similar but exhibit opposite toxicity labels, and construct a novel dataset termed \emph{ToxicityCliff}.

To enable minimal molecular edits, we further decompose each molecule into fragment-level representations using Sequential Attachment-based Fragment Embedding(SAFE) \cite{noutahi2024gotta}. SAFE is a BRICS-based fragment representation that organizes molecules as sequences of chemically meaningful substructures, offering better interpretability than SMILES and aligning naturally with fragment-based drug discovery. Based on this representation, we categorize fragments within each pair into three groups: common fragments shared by both molecules, toxicity-associated fragments present only in the toxic molecule, and non-toxic fragments only in the non-toxic counterpart.
We also compare the frequent toxic fragments with known Structural Alerts in Appendix~\ref{appendix:structural_alert_overlap}.


To improve data quality, we apply an interquartile range (IQR)-based filtering strategy \cite{dash2023outliers} to remove ambiguous instances. For each toxicity cliff pair, we compute the absolute differences between the physicochemical properties of the toxic and non-toxic molecules, and apply IQR filtering to exclude pairs with excessively large property deviations. The considered properties include six descriptors derived from \textit{Lipinski’s 
Rule}~\cite{lipinski1997experimental} and \textit{Veber’s Rule}~\cite{veber2002molecular}, capturing key physicochemical attributes relevant to molecular behavior. As a result, the property distributions of toxic and non-toxic molecules are highly similar, as shown in Figure~\ref{fig:test_property_distribution}, indicating that the physicochemical properties are well preserved within each pair. In addition, we apply IQR-based filtering to remove pairs with unusually large fragment counts. The construction process is illustrated in Figure~\ref{fig:molDetox_motivation}, and detailed filtering procedures are provided in Appendix~\ref{appendix:dataset_construction_details}.

For rigorous evaluation, we partitioned each endpoint using a 9:1 scaffold split, and endpoints with insufficient data were assigned entirely to the test set. Detailed dataset statistics, including the number of training and test samples, are reported in Table~\ref{tab:dataset_statistics_split}.

\subsection{MolDeTox Construction}
A key objective of MolDeTox is to address molecular detoxification as a \emph{hierarchical reasoning process} rather than a simple end-to-end optimization problem. Rather than only asking whether a model can generate a non-toxic molecule, our benchmark explicitly tests whether it can (i) identify the toxicity-relevant fragment, (ii) propose a localized non-toxic replacement, and (iii) integrate such edits into a full non-toxic molecule. This step-by-step formulation makes it possible to diagnose where detoxification succeeds or fails, and to distinguish fragment-level reasoning ability from final generation ability. Examples for each task can be found in Figure~\ref{fig:molDetox_motivation}.

We denote a toxic/non-toxic molecular pair as $(M_t, M_{nt})$, and define the corresponding SAFE-based fragment sets as the shared fragment set $\mathcal{F}_{\mathrm{common}}$, the toxic-only fragment set $\mathcal{F}_{t\text{-only}}$, and the non-toxic-only fragment set $\mathcal{F}_{nt\text{-only}}$, as provided in \emph{ToxicityCliff}. Using these molecule- and fragment-level correspondences, we construct three benchmark tasks that reflect progressively more difficult forms of toxicity-aware reasoning.

\vspace{-0.5em}
\paragraph{Task 1: Toxic Fragment Identification}
Given a toxic molecule $M_t$ represented by its SMILES and the corresponding SAFE fragment set (connected by `.'), the objective is to identify the toxic-only fragment set $\mathcal{F}_{t\text{-only}}$. This task evaluates whether a model can recognize toxicity-relevant substructures by reasoning over both the molecular structure and its fragment-level representation.

\vspace{-0.5em}
\paragraph{Task 2: Non-Toxic Fragment Generation}
The input is the same as in Task 1, with the addition of the toxic-only fragment set $\mathcal{F}_{t\text{-only}}$. The objective is to generate the non-toxic-only SAFE fragment set $\mathcal{F}_{nt\text{-only}}$. This task evaluates whether a model can move beyond identifying problematic substructures and generate plausible non-toxic alternatives at the fragment level.

\vspace{-0.5em}
\paragraph{Task 3: Non-Toxic Molecule Generation}
Similar to Task 1, the input consists of the toxic molecule $M_t$. Unlike Task 2, which focuses on fragment-level replacements, this task requires generating the full non-toxic molecule $M_{nt}$, performing end-to-end detoxification at the whole-molecule level. This task evaluates whether a model can accurately identify toxicity-relevant substructures, apply appropriate modifications, and generate a chemically valid non-toxic molecule.

All three tasks are categorized into single-step and multi-step variants depending on the number of fragments that need to be modified. Detailed descriptions of this setup can be found in Appendix~\ref{appendix:benchmark_construction_details}.
By decomposing the task, MolDeTox offers a granular perspective on a model's capabilities. This allows for a fine-grained analysis of whether the generative failure stems from incorrect toxicity identification or a lack of chemical reasoning during the replacement and integration stages. Moreover, each instance is accompanied by its toxicity endpoint description, providing context that allows the model to adapt to the specific toxicity setting. Endpoint descriptions are provided in Appendix~\ref{appendix:toxicity_endpoint_description}.

\section{Experiments and Results}
\begin{table*}[!t]
\centering
\scriptsize
\setlength{\tabcolsep}{4.0pt}
\renewcommand{\arraystretch}{0.95}
\caption{
Main results on \textbf{Task 1} and \textbf{Task 2} of \textbf{MolDeTox}.
We compare model performance under single-step and multi-step settings.
Best results are in \textbf{bold}, and second-best results are \underline{underlined}.
}
\label{tab:task12_results_clean}

\begin{tabularx}{\textwidth}{lY|YYYY|YYY}
\toprule
& \multicolumn{3}{c}{\textbf{Task 1: Toxic Frag. ID}}
& \multicolumn{5}{c}{\textbf{Task 2: NonToxic Frag. Gen.}} \\
\cmidrule(lr){2-4} \cmidrule(lr){5-9}
\textbf{Model}
& \multicolumn{1}{c|}{\textbf{Single}}
& \multicolumn{2}{c}{\textbf{Multi}}
& \multicolumn{2}{c|}{\textbf{Single}}
& \multicolumn{3}{c}{\textbf{Multi}} \\
& \textbf{Acc.(\%)}
& \textbf{Acc.(\%)} & \textbf{F1}
& \textbf{Acc.(\%)} & \textbf{Lev. Dist}
& \textbf{Acc.(\%)} & \textbf{F1} & \textbf{Lev. Dist} \\
\midrule

\multicolumn{9}{l}{\textit{\textbf{LLMs}}} \\
GPT-4o
& 37.98 & 1.13 & 0.3789
& 5.47 & 4.82 & 0.18 & 0.0141 & 4.87 \\
GPT-5.2
& 41.09 & 3.85 & 0.4358
& \underline{7.92} & \underline{3.96} & 0.23 & 0.0236 & \underline{4.40} \\
GPT-5.2 4-Shot
& \textbf{54.47} & \textbf{13.97} & \underline{0.5573}
& \textbf{20.79} & \textbf{3.64} & \textbf{4.66} & \textbf{0.1234} & \textbf{4.13} \\
Gemini 3.1 Flash-Lite
& 15.63 & 1.05 & 0.1260
& 3.07 & 8.50 & 0.03 & 0.0091 & 6.90 \\
Qwen3-4B-Inst.
& 40.56 & 0.94 & 0.4732
& 0.33 & 4.96 & 0.00 & 0.0005 & 5.78 \\
Qwen3-4B-Inst. 4-Shot
& 36.43 & \underline{7.74} & 0.5134
& 7.75 & 4.64 & \underline{0.65} & \underline{0.0308} & 5.28 \\
Qwen3-8B
& 34.87 & 3.58 & \textbf{0.5648}
& 0.73 & 6.79 & 0.05 & 0.0009 & 4.92 \\
Llama-3.1-8B-Inst.
& 17.51 & 5.54 & 0.4782
& 0.53 & 28.69 & 0.00 & 0.0013 & 40.10 \\
Gemma-3-27B
& \underline{42.45} & 5.79 & 0.4888
& 2.05 & 6.83 & 0.00 & 0.0012 & 8.88 \\
DeepSeek-Llama-70B
& 37.74 & 3.84 & 0.4506
& 3.43 & 5.10 & 0.04 & 0.0092 & 5.97 \\

\addlinespace[2pt]
\multicolumn{9}{l}{\textit{\textbf{VLMs}}} \\
GPT-5.2 w/ Image
& \underline{40.31} & \underline{4.30} & 0.4467
& \textbf{8.15} & \textbf{3.93} & \textbf{0.18} & \textbf{0.0232} & \textbf{4.38} \\
Gemini 3.1 Flash-Lite w/ Image
& \textbf{43.78} & \textbf{4.65} & 0.4361
& \underline{6.69} & \underline{4.40} & \underline{0.08} & \underline{0.0218} & \underline{4.73} \\
Qwen3-VL-4B-Inst.
& 27.73 & 3.50 & \underline{0.5763}
& 0.13 & 5.75 & 0.00 & 0.0005 & 7.08 \\
Qwen3-VL-4B-Inst. w/ Color Image
& 27.78 & 3.06 & \textbf{0.5825}
& 0.10 & 5.57 & 0.00 & 0.0001 & 5.59 \\
Qwen3-VL-8B-Inst.
& 33.46 & 2.01 & 0.5005
& 0.80 & 4.50 & 0.00 & 0.0005 & 6.05 \\
LLaVA-v1.6-Vicuna-13B
& 3.31 & 0.30 & 0.4013
& 0.00 & 12.69 & 0.00 & 0.0014 & 9.19 \\

\bottomrule
\end{tabularx}
\end{table*}
\subsection{Experimental Setup}
We evaluate model performance on the proposed benchmark across three tasks, each under single-step and multi-step settings. For all experiments, we set the sampling temperature to 0.7 and run each model three times, reporting the average performance to ensure robustness. The main tables report the mean performance, while the corresponding standard deviations across the three runs are provided in Appendix~\ref{appendix:result_w_std}. All experiments were conducted using 8 NVIDIA
A100 GPUs.

\paragraph{Prompt Construction}
We design all benchmark prompts with a shared contextual structure to ensure that models interpret each instance under a consistent toxicity-aware setting. Each prompt is composed of four common components: (1) an \textbf{endpoint description} (Table~\ref{tab:endpoint_descriptions}) that specifies the toxicity type, (2) a concise explanation of the \textbf{SAFE representation} (Table~\ref{tab:safe_explanation}) to guide fragment-level reasoning, (3) a \textbf{paired toxic/non-toxic context}(Table~\ref{tab:common_pair_context_prompt}) that explains the relationship between molecule pair in the dataset, and (4) a \textbf{property-preservation instruction}(Table~\ref{tab:property_preservation_prompt}) that emphasizes maintaining physicochemical characteristics during detoxification. These components provide a unified context for all tasks, enabling the model to reason about toxicity and molecular structure in a consistent manner. On top of this shared structure, each task introduces additional instructions tailored to its specific objective. The detailed task-specific prompts are presented in Tables~\ref{tab:task1_question_prompt}--\ref{tab:task3_stepwise_cot_safe_question_prompt}.

\paragraph{Evaluated Models}
We evaluate a diverse set of large language models (LLMs) and vision-language models (VLMs), including both closed-source and open-source models. For LLMs, closed-source models include GPT-4o\cite{hurst2024gpt}, GPT-5.2\cite{openai_gpt5_2_system_card}, and Gemini 3.1 Flash-Lite~\cite{google_gemini_3_1_flash_lite_model_card}, while open-source models include Qwen3-4B-Instruct\cite{yang2025qwen3}, Qwen3-8B, Llama-3.1-8B-Instruct\cite{grattafiori2024llama}, Gemma-3-27B\cite{team2025gemma}, and DeepSeek-Llama-70B\cite{guo2025deepseek}. For VLMs, we evaluate closed-source models such as GPT-5.2 and Gemini 3.1 Flash-Lite with image input, and open-source models including Qwen3-VL-4B-Instruct\cite{bai2025qwen3}, Qwen3-VL-8B-Instruct, and LLaVA-v1.6-Vicuna-13B\cite{liu2023visual}.

\paragraph{Variant Settings}
To explore effective strategies for molecular detoxification, we introduce several variants across the tasks.
First, we apply in-context learning (ICL) by providing four examples (4-shot) retrieved from the training set of the same dataset endpoint based on structural similarity to the query molecule, inspired by the retrieval strategy in MolRAG\cite{xian2025molrag}. If no training examples are available for a given endpoint, inference is performed without ICL.
For VLMs, we augment the input by providing fragment-aware molecular images, where SAFE fragments are highlighted with distinct colors (w/ Color Image).
In Task 3, we additionally investigate a step-by-step reasoning approach (w/ CoT), where the model first performs toxic fragment identification(Task 1) and replacement(Task 2) before generating the final non-toxic molecule. The corresponding prompt example is provided in Table~\ref{tab:task3_stepwise_cot_safe_system_prompt}.
Finally, instead of directly generating SMILES strings (SMILES Generation), we propose a SAFE-based generation strategy (SAFE Generation), where the model generates SAFE strings, which are then converted into a valid molecule using a SAFE-to-SMILES decoding function\footnote{\url{https://github.com/datamol-io/safe}}.
\vspace{-0.5em}

\begin{table*}[!t]
\centering
\scriptsize
\setlength{\tabcolsep}{4.2pt}
\renewcommand{\arraystretch}{0.95}
\caption{
Comparison of inference strategies on \textbf{Task 3} of \textbf{MolDeTox}.
We compare model performance under single-step and multi-step settings across SMILES and SAFE generation.
Best results are in \textbf{bold}, and second-best results are \underline{underlined}.
}
\label{tab:task3_results}

\resizebox{\textwidth}{!}{%
\begin{tabular}{lcccccccc}
\toprule
\textbf{Model} & \textbf{Acc.(\%)} & \textbf{BLEU1} & \textbf{Levenshtein} & \textbf{RDK FTS} & \textbf{MACCS FTS} & \textbf{Morgan FTS} & \textbf{Validity} & \textbf{PRS} \\
\midrule

\rowcolor{gray!12}
\multicolumn{9}{c}{\textbf{Single Step}} \\

\multicolumn{9}{l}{\textit{\textbf{SMILES Generation}}} \\
\multicolumn{9}{l}{\textit{\textbf{LLMs}}} \\
GPT-5.2            & \textbf{3.50}    & \underline{0.9411} & \textbf{7.29}     & 0.7020             & \underline{0.7532} & 0.5575             & \textbf{0.9296}    & \underline{0.7700} \\
Qwen3-4B-Inst.     & 0.13             & 0.8098             & \underline{22.03} & \textbf{0.7939}    & \textbf{0.7957}    & \textbf{0.6484}    & \underline{0.9176} & \textbf{0.7755} \\
Qwen3-8B           & 0.40             & \textbf{0.9592}    & 35.07             & \underline{0.7297} & 0.7094             & \underline{0.5924} & 0.8475             & 0.6549 \\
DeepSeek-Llama-70B & \underline{1.88} & 0.8976             & 556.68            & 0.4634             & 0.4948             & 0.3578             & 0.7027             & 0.5482 \\

\addlinespace[2pt]
\multicolumn{9}{l}{\textit{\textbf{SAFE Generation}}} \\
\multicolumn{9}{l}{\textit{\textbf{LLMs}}} \\
GPT-4o                     & 1.98             & 0.8528             & 8.22              & 0.6703             & 0.7237             & 0.5358             & 0.9154             & 0.7246 \\
GPT-5.2                    & \underline{4.00} & 0.8827             & 6.70              & 0.7198             & 0.7625             & 0.5825             & 0.9362             & 0.7408 \\
GPT-5.2 w/ CoT             & 3.77             & 0.8847             & 6.63              & 0.7108             & 0.7543             & 0.5768             & 0.9402             & 0.7438 \\
GPT-5.2 4-Shot             & \textbf{15.59}   & 0.9140             & 6.44              & 0.7768             & 0.8155             & 0.6517             & 0.9690             & \underline{0.9224} \\
Gemini 3.1 Flash-Lite      & 0.43             & 0.6046             & 29.94             & 0.3565             & 0.3463             & 0.1538             & 0.8866             & 0.6906 \\
Qwen3-4B-Inst.             & 0.07             & \textbf{0.9535}    & \underline{3.50}  & \textbf{0.8600}    & \textbf{0.8675}    & \textbf{0.7094}    & \textbf{0.9983}    & 0.8170 \\
Qwen3-4B-Inst. w/ CoT      & 0.43             & 0.7845             & 9.75              & 0.6436             & 0.6866             & 0.4941             & 0.8618             & 0.6752 \\
Qwen3-4B-Inst. 4-Shot      & 3.38             & \underline{0.9418} & 5.18              & \underline{0.8217} & \underline{0.8651} & \underline{0.6991} & \underline{0.9944} & \textbf{0.9815} \\
Qwen3-8B                   & 0.00             & 0.6432             & \textbf{1.95}     & 0.5920             & 0.5839             & 0.4818             & 0.6703             & 0.8319 \\
Llama-3.1-8B-Inst.         & 0.00             & 0.2904             & 19.00             & 0.0914             & 0.0903             & 0.0515             & 0.5325             & 0.5850 \\
Gemma-3-27B                & 1.55             & 0.8532             & 7.39              & 0.7313             & 0.7792             & 0.5963             & 0.9419             & 0.7327 \\
DeepSeek-Llama-70B         & 2.30             & 0.7124             & 11.10             & 0.5189             & 0.5548             & 0.3922             & 0.7804             & 0.6140 \\

\addlinespace[2pt]
\multicolumn{9}{l}{\textit{\textbf{VLMs}}} \\
GPT-5.2 w/ Image                 & \underline{2.61}    & 0.8938             & 7.49              & 0.7330             & 0.7665             & 0.5872             & 0.9452             & 0.7652 \\
Gemini 3.1 Flash-Lite w/ Image   & \textbf{3.19}    & 0.8228             & 9.68              & 0.6600             & 0.7018             & 0.5079             & 0.8879             & 0.6929 \\
Qwen3-VL-4B-Inst.                & 0.07             & \underline{0.9288} & \underline{4.74}  & \underline{0.7895} & \underline{0.8426} & \underline{0.6705} & \underline{0.9835} & \underline{0.7778} \\
Qwen3-VL-4B-Inst. w/ Color Image & 0.10             & \textbf{0.9364}    & \textbf{4.08}     & \textbf{0.8199}    & \textbf{0.8491}    & \textbf{0.6849}    & \textbf{0.9866}    & \textbf{0.7931} \\
Qwen3-VL-8B-Inst.                & 0.50 & 0.9076             & 6.53              & 0.7763             & 0.8130             & 0.6375             & 0.9663             & 0.7765 \\
LLaVA-v1.6-Vicuna-13B            & 0.00             & 0.3559             & 18.58             & 0.1210             & 0.1543             & 0.0814             & 0.5380             & 0.3912 \\

\midrule
\rowcolor{gray!12}
\multicolumn{9}{c}{\textbf{Multi Step}} \\

\multicolumn{9}{l}{\textit{\textbf{SMILES Generation}}} \\
\multicolumn{9}{l}{\textit{\textbf{LLMs}}} \\
GPT-5.2            & 0.21             & \underline{0.8940} & \underline{13.17} & 0.5875             & \textbf{0.6278}    & 0.4084             & \textbf{0.9327}    & \textbf{0.7213} \\
Qwen3-4B-Inst.     & 0.00             & 0.8624             & 15.30             & \textbf{0.6401}    & \underline{0.6249} & \textbf{0.4445}    & 0.8429             & 0.5918 \\
Qwen3-8B           & \underline{0.28} & \textbf{0.9407}    & \textbf{11.99}    & \underline{0.6175} & 0.6171             & \underline{0.4183} & \underline{0.8560} & \underline{0.6017} \\
DeepSeek-Llama-70B & \textbf{0.60}    & 0.8619             & 530.28            & 0.3939             & 0.4141             & 0.2654             & 0.6825             & 0.5195 \\

\addlinespace[2pt]
\multicolumn{9}{l}{\textit{\textbf{SAFE Generation}}} \\
\multicolumn{9}{l}{\textit{\textbf{LLMs}}} \\
GPT-4o                     & 0.10             & 0.8138             & 13.11             & 0.5636             & 0.6033             & 0.3854             & 0.9222             & 0.7143 \\
GPT-5.2                    & 0.15             & 0.8553             & 11.28             & 0.6109             & 0.6503             & 0.4352             & 0.9525             & 0.7381 \\
GPT-5.2 w/ CoT             & 0.23             & 0.8418             & 11.60             & 0.5932             & 0.6355             & 0.4192             & 0.9417             & 0.7289 \\
GPT-5.2 4-Shot             & \textbf{2.95}    & 0.8604             & 11.60             & 0.6469             & 0.6831             & 0.4779             & 0.9558             & \underline{0.8987} \\
Gemini 3.1 Flash-Lite      & 0.21             & 0.7182             & 23.27             & 0.4392             & 0.4332             & 0.2355             & 0.9346             & 0.7412 \\
Qwen3-4B-Inst.             & 0.00             & \textbf{0.9409}    & \underline{10.27} & \textbf{0.7546}    & \textbf{0.7871}    & \textbf{0.5282}    & \underline{0.9974} & 0.7436 \\
Qwen3-4B-Inst. w/ CoT      & 0.03             & 0.8225             & 13.16             & 0.5912             & 0.6443             & 0.4024             & 0.9095             & 0.6559 \\
Qwen3-4B-Inst. 4-Shot      & \underline{0.49} & \underline{0.8968} & 10.92             & \underline{0.6917} & \underline{0.7324} & \underline{0.5194} & \textbf{0.9992}    & \textbf{0.9854} \\
Qwen3-8B                   & 0.00             & 0.6250             & \textbf{7.85}     & 0.4927             & 0.5036             & 0.3480             & 0.6682             & 0.7342 \\
Llama-3.1-8B-Inst.         & 0.00             & 0.2895             & 17.95             & 0.0903             & 0.0952             & 0.0535             & 0.5465             & 0.4472 \\
Gemma-3-27B                & 0.05             & 0.7972             & 13.24             & 0.5946             & 0.6352             & 0.4167             & 0.9332             & 0.7114 \\
DeepSeek-Llama-70B         & 0.26             & 0.6964             & 12.85             & 0.4591             & 0.4779             & 0.2989             & 0.7891             & 0.6097 \\

\addlinespace[2pt]
\multicolumn{9}{l}{\textit{\textbf{VLMs}}} \\
GPT-5.2 w/ Image                 & 0.15             & \underline{0.8890} & \underline{10.53} & \underline{0.6483} & \underline{0.6861} & 0.4452             & 0.9592 & 0.7224 \\
Gemini 3.1 Flash-Lite w/ Image   & \underline{0.21} & 0.7975             & 13.89             & 0.5709             & 0.6043             & 0.3836             & 0.9138             & 0.7273 \\
Qwen3-VL-4B-Inst.                & 0.03             & 0.8680             & 11.67             & 0.6357             & 0.6785             & \underline{0.4638} & \underline{0.9837} & \textbf{0.7549} \\
Qwen3-VL-4B-Inst. w/ Color Image & 0.00             & \textbf{0.8944}    & \textbf{10.49}    & \textbf{0.6914}    & \textbf{0.7253}    & \textbf{0.5088}    & \textbf{0.9888}    & \underline{0.7502} \\
Qwen3-VL-8B-Inst.                & \textbf{0.41}    & 0.7901             & 12.45             & 0.5564             & 0.5834             & 0.3925             & 0.9070             & 0.7148 \\
LLaVA-v1.6-Vicuna-13B            & 0.00             & 0.2984             & 13.88             & 0.1274             & 0.1510             & 0.0854             & 0.4870             & 0.3340 \\

\bottomrule
\end{tabular}%
}
\end{table*}

\subsection{Evaluation Metrics}
We adopt task-specific evaluation metrics to comprehensively assess model performance across the three tasks.
For Task 1, we evaluate whether the predicted fragment exactly matches the ground-truth fragment using accuracy. For the multi-fragment setting, where multiple fragments are predicted, we report the F1 score to account for partial correctness.
For Task 2, we report accuracy for the single-setting, and both accuracy and F1 for the multi-setting. In addition, we measure the Levenshtein distance to quantify the lexical similarity between generated fragments and ground-truth fragments. It captures how closely the generated fragments resemble valid non-toxic alternatives.
For Task 3, we adopt standard molecule generation metrics widely used in prior work\cite{yang2025knowmol}, including validity, similarity-based metrics, and string matching, following established evaluation protocols.
Furthermore, we introduce a Property Retention Score (PRS) to quantify how well the generated non-toxic molecule preserves the physicochemical properties of the original toxic molecule. Inspired by the Quantitative Estimate of Drug-likeness (QED), we compute a weighted average over a subset of physicochemical property functions derived from \textit{Lipinski’s Rule} \cite{lipinski1997experimental} and \textit{Veber’s Rule} \cite{veber2002molecular}. We then measure the absolute difference between the scores of the toxic molecule and the generated molecule, and transform this difference into a normalized score in the range [0, 1]. Smaller differences yield scores closer to 1, indicating better property preservation. Detailed descriptions of all evaluation metrics are provided in Appendix~\ref{appendix:evaluation_metrics}.




\subsection{Performance on MolDeTox across Models}
Tables~\ref{tab:task12_results_clean},~\ref{tab:task3_results} show a consistent hierarchy of difficulty across all models. Performance is highest on Task 1, decreases on Task 2, and drops further on Task 3. This pattern suggests that the tasks become progressively more challenging as they require less explicit guidance and greater generative flexibility, leading to increased difficulty for current models. Despite these challenges, in-context learning is the most effective strategy overall. In particular, 4-shot consistently improves performance across tasks, indicating that models rely heavily on explicit examples to learn fragment-level editing patterns. In contrast, CoT prompting does not consistently help, especially in multi-step settings where errors from earlier steps affect the final output. 

Specifically for Task 3, we compare direct SMILES generation and SAFE generation within the same models. Aggregated across single-step and multi-step settings on this common set of models, SAFE generation performs better than direct SMILES generation on RDK (+0.0616), MACCS (+0.0632), and Morgan FTS (+0.0492), as well as on Validity (+0.0690) and PRS (+0.0555), with all differences significant under a paired $t$-test ($p<0.001$). These results suggest that representing molecules as explicit fragment compositions is better suited to localized detoxification. It allows models to modify toxicity-relevant substructures while more reliably preserving the remaining molecular structure and physicochemical properties than direct SMILES generation

VLMs follow similar trends to LLMs. Comparing GPT-5.2 in both LLM and VLM settings, there is no large performance gap on Task 1 and Task 2, indicating that visual inputs do not significantly change performance in fragment identification or replacement. On Task 3 (single-step), the LLM achieves higher accuracy, while the VLM shows better structural validity and property retention (PRS), suggesting a trade-off between exact matching and chemically plausible generation.
A notable pattern is observed in the Qwen3-VL series. Larger models consistently achieve higher accuracy, indicating that scaling improves performance even in multimodal settings. In addition, when color-enhanced images highlighting fragment-level regions are provided, both single-step and multi-step settings show improved generation quality across most metrics except accuracy. This suggests that visual emphasis helps models better preserve local structural consistency, even if it does not directly improve exact molecule reconstruction. 
We also report endpoint-wise results in 
Appendix~\ref{appendix:endpoint_wise_result_analysis}.

\section{Task-wise Success Dependency Analysis}
\label{task_wise_success_analysis}

\begin{figure}[h]
    \centering

    \begin{minipage}[t]{0.32\columnwidth}
        \centering
        \vspace{0.35em}
        \scriptsize
        \setlength{\tabcolsep}{2.5pt}
        \renewcommand{\arraystretch}{1.15}
    
        \scalebox{1.08}{%
        \begin{tabular}{c|c}
        \hline
        \textbf{Case} & \textbf{Pattern} \\
        \hline
        C000 & $T1{=}0,T2{=}0,T3{=}0$ \\
        C100 & $T1{=}1,T2{=}0,T3{=}0$ \\
        C010 & $T1{=}0,T2{=}1,T3{=}0$ \\
        C110 & $T1{=}1,T2{=}1,T3{=}0$ \\
        C001 & $T1{=}0,T2{=}0,T3{=}1$ \\
        C101 & $T1{=}1,T2{=}0,T3{=}1$ \\
        C011 & $T1{=}0,T2{=}1,T3{=}1$ \\
        C111 & $T1{=}1,T2{=}1,T3{=}1$ \\
        \hline
        \end{tabular}
        }

        \vspace{-2pt}
        \captionsetup{type=table, font=small, justification=centering, singlelinecheck=true}
        \captionof{table}{Eight outcome cases for step-wise dependency analysis.}
        \label{tab:stepwise_case_definition}
    \end{minipage}%
    \hspace{0.01\columnwidth}%
    \begin{minipage}[t]{0.64\columnwidth}
        \centering
        \vspace{0pt}
        \includegraphics[width=\linewidth]{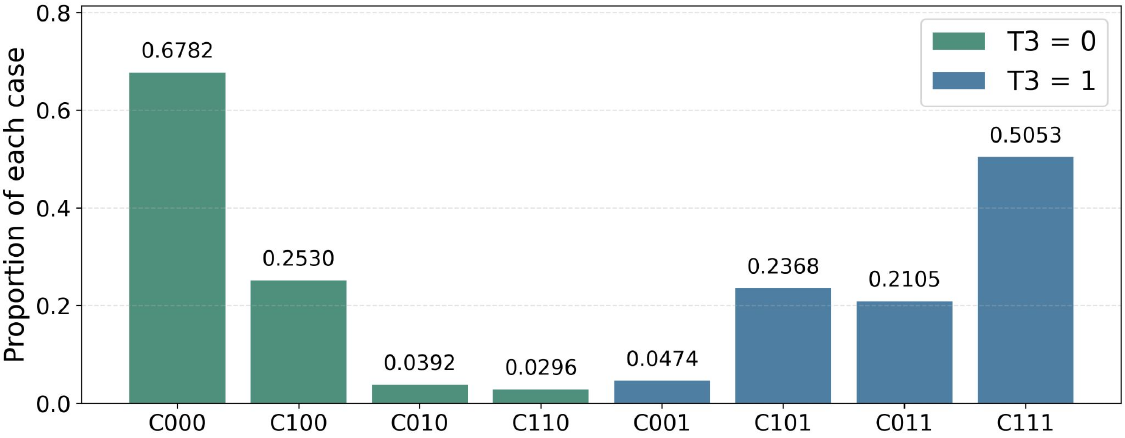}

        \vspace{-2pt}
        \captionsetup{type=figure, font=small, justification=centering, singlelinecheck=true}
        \captionof{figure}{GPT-5.2 4-shot $T3$-conditioned outcome proportion.}
        \label{fig:step_wise_barplot}
    \end{minipage}

    \vspace{-6pt}
\end{figure}

To better understand how models achieve final success or failure, we categorize each test sample into one of eight outcome cases defined by the correctness patterns of Task~1, Task~2, and Task~3. Table~\ref{tab:stepwise_case_definition} presents the case definitions, and Figure~\ref{fig:step_wise_barplot} visualizes the conditional probability distributions for GPT-5.2 under the 4-shot setting, separated into final success ($T3=1$) and final failure ($T3=0$).

A clear pattern in the distributions is that Task~3 failure is dominated by C000. Within the $T3=0$ group, C000 accounts for 0.6782, far exceeding all other failure-side cases, which indicates that most failed samples reflect a complete breakdown across all three tasks rather than a near-miss at the final stage. The next largest failure side case is C100 (0.2530), suggesting that models can often identify the toxic fragment correctly but still fail to carry this partial success into replacement and final molecule generation. By contrast, C010 and C110 are much less frequent, at 0.0392 and 0.0296.

\begin{wrapfigure}{r}{0.65\textwidth}
    \centering
    \vspace{-1.0em}
    \includegraphics[width=0.63\textwidth]{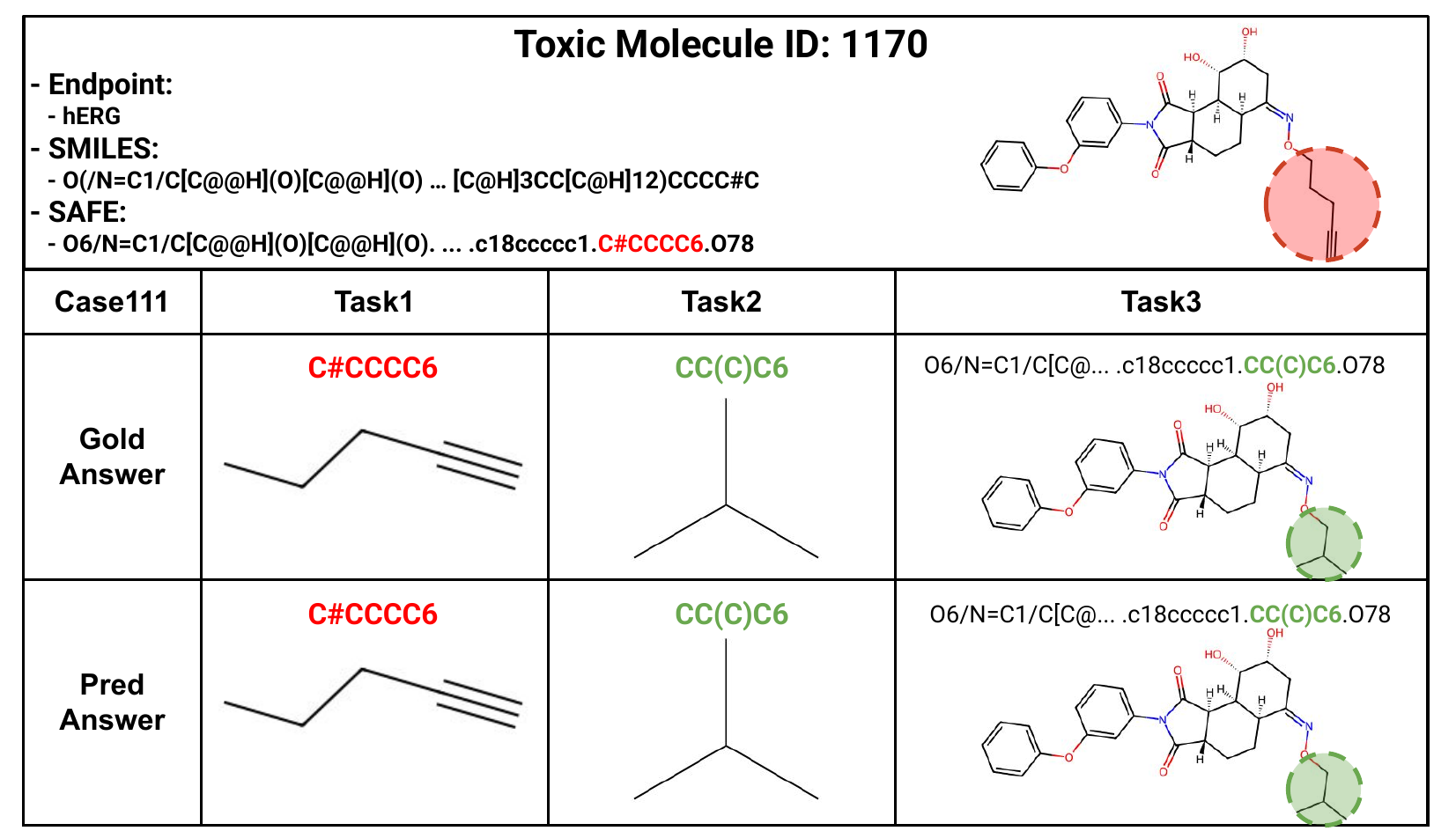}
    \caption{Example of Pattern $T1{=}1,T2{=}1,T3{=}1$ (C111).}
    \label{fig:c111_case_study}
    \vspace{-1.0em}
\end{wrapfigure}

In successful cases, final outcomes are typically accompanied by at least partially correct intermediate steps. These samples are concentrated in C111 (0.5053) and C101 (0.2368), showing that Task~3 success most often arises either from fully correct step-wise execution or from cases where correct toxic-fragment identification alone is sufficient to support the final generation. A C111 example is shown in Figure~\ref{fig:c111_case_study}. C011 also occupies a substantial portion of the success group (0.2105), indicating that correct replacement can still support final success even when the initial toxic-fragment identification is incorrect. In contrast, C001 is rare (0.0474), suggesting that final-only success without any correct intermediate step is uncommon.

Taken together, these results show that the three-step decomposition in MolDeTox provides a meaningful unit of analysis rather than a purely auxiliary task design. Final success typically occurs when all intermediate steps are correct. In contrast, failures arise either when all steps break down, or when correct toxic-fragment identification does not carry over to the subsequent replacement and final generation stages. This structure makes MolDeTox an interpretable benchmark for analyzing not only whether detoxification succeeds, but also where and how the step-wise process breaks down. Appendix~\ref{appendix:case_examples} provides examples for the remaining seven outcome cases.
\section{Conclusion and Discussion}
In this work, we introduced \textbf{MolDeTox}, a benchmark for evaluating toxicity-aware molecular optimization through step-wise reasoning. Built on a \emph{ToxicityCliff} dataset of structurally similar molecular pairs with opposite toxicity labels, MolDeTox formulates detoxification as a minimal-edit process, requiring models to identify toxic fragments, propose localized edits, and generate non-toxic molecules while preserving the original structure. We further incorporate a fragment-level representation based on SAFE to enable more interpretable and localized molecular editing.

Our experiments on LLMs and VLMs show that molecular detoxification remains challenging, with performance degrading significantly from fragment-level identification to full molecule generation. We find that in-context learning and SAFE-based generation substantially improve performance, while step-by-step reasoning approaches are limited by error propagation. This challenge is further amplified by our evaluation protocol, which avoids proxy toxicity predictors and instead compares generated molecules against ground-truth non-toxic counterparts derived from real data. As a result, the task becomes inherently more difficult, leading to lower absolute performance. Rather, MolDeTox provides a structured framework for analyzing step-wise molecular editing and offers insights for developing more reliable and interpretable toxicity-aware molecular design methods.

Despite these advances, we observe notable performance variations across tasks, suggesting that current models are not equally effective at all stages. As future work, we plan to develop task-specialized models via supervised fine-tuning or reasoning-aware reinforcement learning. We also aim to integrate these models into an agentic framework for more effective end-to-end detoxification. In real-world drug discovery, promising candidates are often discarded due to toxicity despite high efficacy. Such approaches could enable targeted detoxification, allowing these compounds to be refined rather than abandoned while reducing safety risks.

\section{Acknowledgments}
This research was supported by (1) the National Research Foundation of Korea (NRF-2023R1A2C3004176), (2) the Ministry of Health \& Welfare, Republic of Korea (HR20C002103), (3) ICT Creative Consilience Program through the Institute of Information \& Communications Technology Planning \& Evaluation (IITP) grant funded by the Korea government (MSIT) (IITP-2026-RS-2020-II201819), (4) the National Research Foundation of Korea (NRF) grant funded by the Korea government (MSIT and MOE) (No. RS-2025-16652968), (5) the Seoul National University Hospital with support from the Ministry of Science and ICT (RS-2023-00262002) and (6) the Korea Bio Data Station (K-BDS) with computing resources including technical support.


\bibliographystyle{unsrtnat}

\clearpage
\appendix
\renewcommand{\thetable}{\Alph{table}}
\setcounter{table}{0}
\section{\emph{\textbf{ToxicityCliff}} Construction Details}
\label{appendix:dataset_construction_details}
\begin{table*}[!h]
\centering
\footnotesize
\setlength{\tabcolsep}{4.2pt}
\renewcommand{\arraystretch}{1.12}
\caption{Statistics of \textbf{\emph{ToxicityCliff}}, including the number of endpoints, molecules, toxicity cliff pairs, and resulting split sizes across datasets.}
\label{tab:dataset_statistics_split}
\begin{adjustbox}{max width=\textwidth}
\begin{tabular}{lrrrrrrrrrrr}
\toprule
\textbf{Dataset}
& \textbf{DILIst}
& \textbf{DICTrank}
& \textbf{DIRIL}
& \textbf{hERG}
& \textbf{AMES}
& \textbf{Skin Reaction}
& \textbf{Tox21}
& \textbf{ClinTox}
& \textbf{CYP Inh.}
& \textbf{SIDER}
& \textbf{Total} \\
\midrule
\textbf{Endpoint N} & 1 & 1 & 1 & 1 & 1 & 1 & 12 & 1 & 5 & 25 & 49 \\
\textbf{Molecule N} & 97 & 27 & 6 & 4,303 & 1,133 & 56 & 1,635 & 30 & 3,355 & 212 & 10,854 \\
\textbf{\emph{ToxicityCliff} N} & 93 & 20 & 3 & 5,355 & 6,444 & 130 & 30,370 & 15 & 8,502 & 1,953 & 52,885 \\
\addlinespace[2pt]
\textbf{Train} & 84 & 18 & 0 & 4,831 & 6,325 & 124 & 29,946 & 14 & 7,807 & 1,434 & 50,583 \\
\textbf{Test} & 9 & 2 & 3 & 524 & 119 & 6 & 424 & 1 & 695 & 519 & 2,302 \\
\bottomrule
\end{tabular}
\end{adjustbox}
\end{table*}

To construct MolDeTox, we design a multi-stage pairing pipeline that identifies toxic/non-toxic molecule pairs that differ in toxicity label while remaining highly similar in overall structure and molecular characteristics. Our goal is to frame detoxification as a \emph{localized molecular editing problem}, where toxicity changes can be attributed to a small number of fragment-level modifications rather than a complete molecular redesign.

Let $M_t$ and $M_{nt}$ denote a toxic molecule and a non-toxic molecule, respectively. The final toxicity cliff pairs are constructed through the following steps.

\paragraph{Step 1: Toxic/Non-Toxic candidate pairing}
For each toxicity dataset, we first split molecules according to the endpoint label and canonicalize all SMILES strings. We then construct candidate toxic/non-toxic pairs using stringent whole-molecule similarity criteria, as in MoleculeACE~\cite{van2022exposing}. A pair $(M_t, M_{nt})$ is retained if it satisfies \emph{at least one} of the following:

\begin{itemize}
    \item Bemis--Murcko scaffold-based ECFP4 Tanimoto similarity $\geq 0.9$,
    \item Full-SMILES ECFP4 Tanimoto similarity $\geq 0.9$,
    \item Normalized SMILES Levenshtein similarity $\geq 0.9$
\end{itemize}

These criteria ensure that paired molecules are globally similar enough to support meaningful toxicity-aware editing.

\paragraph{Step 2: SAFE conversion}
To analyze local structural differences, we convert each paired SMILES into its SAFE representation. SAFE expresses a molecule as a dot-separated sequence of fragments, allowing explicit comparison of shared and differing substructures between toxic and non-toxic molecules.

Formally, let
\[
S_t = \mathrm{SAFE}(M_t), \qquad S_{nt} = \mathrm{SAFE}(M_{nt})
\]
where each SAFE string is decomposed into fragment sets by splitting on the dot separator:
\[
\mathcal{F}_t = \mathrm{split}(S_t, \texttt{`.'}), \qquad
\mathcal{F}_{nt} = \mathrm{split}(S_{nt}, \texttt{`.'})
\]

\paragraph{Step 3: SAFE fragment comparison}
Using the fragment sets $\mathcal{F}_t$ and $\mathcal{F}_{nt}$, we define three fragment groups:

\[
\mathcal{F}_{\mathrm{common}} = \mathcal{F}_t \cap \mathcal{F}_{nt},
\]
\[
\mathcal{F}_{t\text{-only}} = \mathcal{F}_t \setminus \mathcal{F}_{nt},
\]
\[
\mathcal{F}_{nt\text{-only}} = \mathcal{F}_{nt} \setminus \mathcal{F}_t.
\]

We also record their cardinalities:
\[
n_{\mathrm{common}} = |\mathcal{F}_{\mathrm{common}}|,\quad
n_{t\text{-only}} = |\mathcal{F}_{t\text{-only}}|,\quad
n_{nt\text{-only}} = |\mathcal{F}_{nt\text{-only}}|
\]

This representation provides a fragment-level view of which substructures are preserved and which are edited between the toxic and non-toxic molecules.

\paragraph{Step 4: SAFE-based fragment filtering}
We next filter candidate pairs to retain only those that reflect small and interpretable fragment-level edits. Concretely, we apply the following rules:

\begin{itemize}
    \item \textbf{Shared core constraint:}
    \[
    n_{\mathrm{common}} \neq 0
    \]
    This ensures that the toxic and non-toxic molecules preserve at least one common fragment.

    \item \textbf{Non-trivial edit constraint:}
    pairs with
    \[
    n_{t\text{-only}} = 0 \quad \text{and} \quad n_{nt\text{-only}} = 0
    \]
    are removed, since they do not differ at the fragment level.

    \item \textbf{Fragment length outlier filtering:}
    we remove pairs if any fragment in $\mathcal{F}_{t\text{-only}}$ or $\mathcal{F}_{nt\text{-only}}$ has SAFE length above an outlier threshold determined from the empirical fragment-length distribution using an interquartile-range (IQR)-based rule. In our benchmark, this corresponds to removing pairs containing toxic-only or non-toxic-only fragments with SAFE length $\geq 28$.

    \item \textbf{Fragment count outlier filtering:}
    we restrict the number of edited fragments using thresholds derived from the empirical distributions of $n_{t\text{-only}}$ and $n_{nt\text{-only}}$ under the same IQR-based outlier rule. In practice, we retain only pairs satisfying
    \[
    n_{t\text{-only}} \leq 4
    \qquad \text{and} \qquad
    n_{nt\text{-only}} \leq 4
    \]
    This removes cases requiring excessively many fragment edits.
\end{itemize}

Together, these rules ensure that the remaining pairs reflect localized, minimal, and interpretable fragment-level differences.

\paragraph{Step 5: Molecular property filtering}
To ensure that paired molecules differ mainly in toxicity rather than in broad physicochemical characteristics, we compute RDKit descriptors for both molecules:
\[
\{\mathrm{MW}, \mathrm{logP}, \mathrm{TPSA}, \mathrm{HBD}, \mathrm{HBA}, \mathrm{RotB}\}
\]

For each descriptor $p$, we compute the absolute difference
\[
\Delta p = |p(M_t) - p(M_{nt})|
\]

We then remove pairs that exhibit outlier-level changes in any descriptor using an interquartile-range (IQR)-based filtering procedure over the empirical distribution of descriptor differences. That is, if a pair is an outlier in at least one molecular property difference, it is discarded. This step reduces unrealistic or pharmacologically mismatched pairs and preserves drug-like similarity beyond fragment-level overlap.

\paragraph{Step 6: Final toxicity cliff pairs}
The remaining pairs form the final \textbf{toxicity cliff pairs} used in MolDeTox. Each retained pair is characterized by:
\begin{itemize}
    \item \textbf{high global similarity} at the whole-molecule level,
    \item \textbf{localized and interpretable fragment differences} under SAFE decomposition,
    \item \textbf{limited physicochemical deviation} across key molecular properties
\end{itemize}

As a result, MolDeTox focuses on toxic/non-toxic pairs where detoxification can be interpreted as a small number of meaningful fragment edits, making the benchmark well-suited for evaluating whether models can identify toxicity-relevant fragments, propose minimal non-toxic edits, and generate full non-toxic analogs while preserving the key characteristics of the original molecule.

\newpage
\section{Comparative analysis of Structural Alerts}
\label{appendix:structural_alert_overlap}

\begin{figure*}[h]
    \centering
    \includegraphics[width=\textwidth]{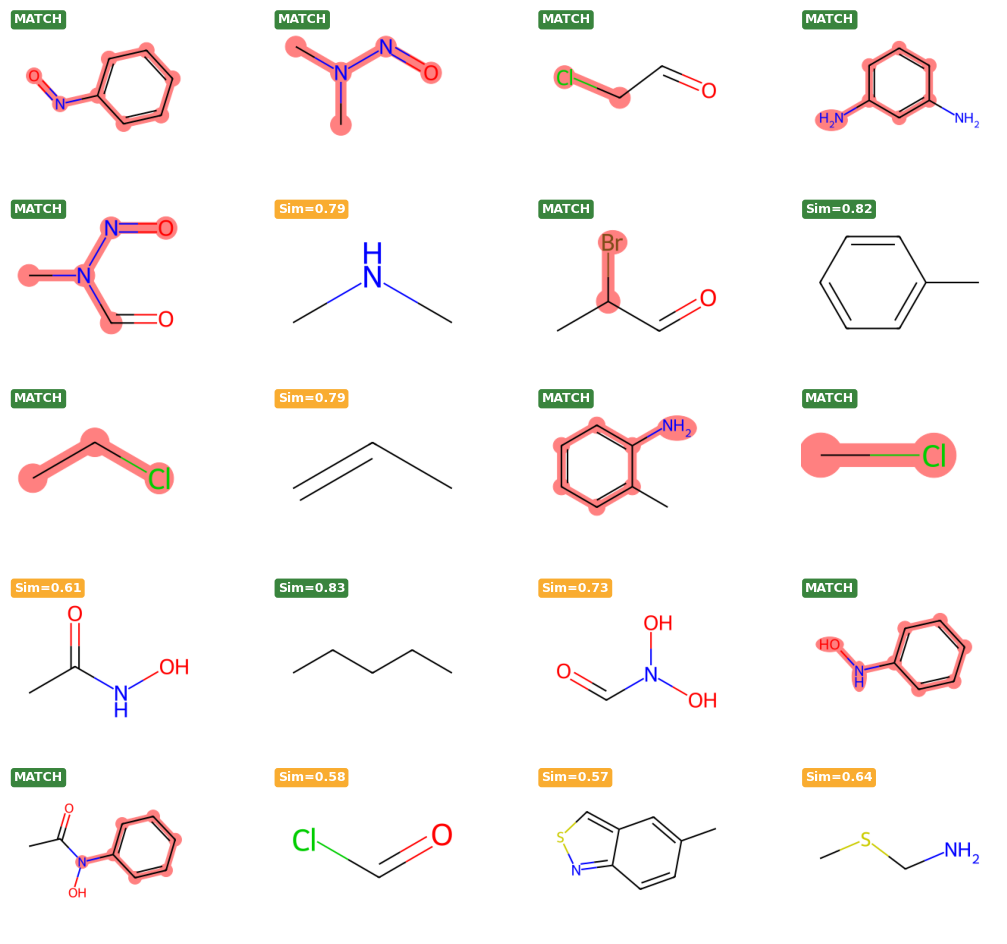}
    \caption{Structural alert overlap for the top-20 Ames Mutagenicity toxicity-associated fragments.}
    \label{fig:ames}
\end{figure*}

\begin{figure*}[h]
    \centering
    \includegraphics[width=\textwidth]{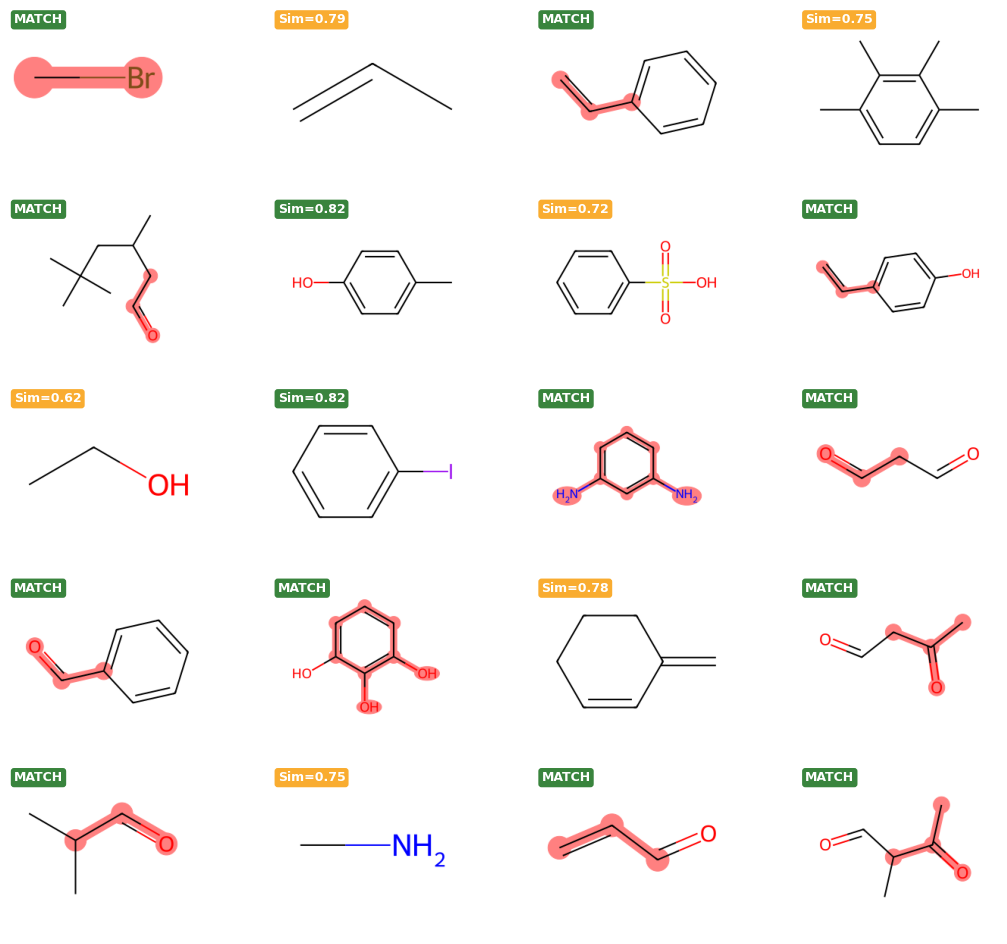}
    \caption{Structural alert overlap for the top-20 Skin Reaction toxicity-associated fragments.}
    \label{fig:skin_reaction}
\end{figure*}

To assess the reliability of extracted toxic fragments, we further compare them against known structural alerts (SAs) collected in ToxAlerts~\cite{sushko2012toxalerts}. ToxAlerts is a curated repository of toxicity-related SAs collected from expert knowledge and literature sources for various toxicity endpoints. We select the most closely aligned endpoints, namely Ames Mutagenicity and Skin Reaction, and compare the top-20 most frequent toxicity-associated fragments from each dataset against the corresponding endpoint-specific SAs.

For each extracted fragment, we first determine whether it directly matches a known SA for the corresponding endpoint. For fragments without an exact match, we compute the maximum Tanimoto similarity between the fragment and the full set of SAs using 1024-bit ECFP4 fingerprints. In Figures~\ref{fig:ames} and~\ref{fig:skin_reaction}, green labels indicate either exact SA matches or high structural similarity scores of at least 0.8, while yellow \texttt{Sim=} labels indicate moderate structural similarity scores of at least 0.5.

A substantial portion of the extracted fragments directly matches known SAs, while many of the remaining fragments still exhibit moderate structural similarity to documented alerts. These results demonstrate that the toxic fragments extracted by MolDeTox are strongly aligned with established toxicity knowledge and capture meaningful toxicity-relevant substructures. This supports the reliability of our data-driven fragment extraction procedure and suggests that the toxic-only fragments derived from \emph{ToxicityCliff} provide chemically grounded targets for toxicity-aware molecular editing.

\newpage
\section{MolDeTox Construction Details}
\label{appendix:benchmark_construction_details}

\subsection{Task Construction Details}
MolDeTox instances are constructed from toxicity cliff pairs $(M_t, M_{nt})$ and their corresponding SAFE fragment decompositions. From each pair, we derive the shared fragment set $\mathcal{F}_{\mathrm{common}}$, the toxic-only fragment set $\mathcal{F}_{t\text{-only}}$, and the non-toxic-only fragment set $\mathcal{F}_{nt\text{-only}}$, which are then used to instantiate the three benchmark tasks.

\paragraph{Task 1: Toxic Fragment Identification}
For Task~1, the input is the toxic molecule $M_t$, and the target output is the toxic-only SAFE fragment set $\mathcal{F}_{t\text{-only}}$.

\paragraph{Task 2: Non-Toxic Fragment Generation}
For Task~2, the input consists of the toxic molecule $M_t$ together with the toxic-only fragment set $\mathcal{F}_{t\text{-only}}$, and the target output is the non-toxic-only SAFE fragment set $\mathcal{F}_{nt\text{-only}}$.

\paragraph{Task 3: Non-Toxic Molecule Generation}
For Task~3, the input is the toxic molecule $M_t$, and the target output is the full non-toxic molecule $M_{nt}$.

\paragraph{Single-step and Multi-step settings}
We divide benchmark instances into single-step and multi-step settings based on the number of fragment tokens in the SAFE labels. Let $n_t$ and $n_{nt}$ denote the numbers of toxic-only and non-toxic-only fragments, respectively.

For Task~1, an instance is single-step if $n_t = 1$, and multi-step if $n_t \ge 2$.
For Task~2 and Task~3, an instance is single-step if $n_t = 1$ and $n_{nt} = 1$, and multi-step if $n_t \ge 2$ or $n_{nt} \ge 2$.
This partition is applied consistently to all QA variants built from the same molecule-pair rows.

\subsection{Benchmark Statistics}

The raw toxicity cliff pair statistics and endpoint-level split results are summarized in the main paper in Table~\ref{tab:dataset_statistics_split}. In total, MolDeTox is constructed from 52,885 toxicity cliff pairs derived from 10,854 molecules across 49 toxicity endpoints.

Building on these toxicity cliff pairs, Table~\ref{tab:appendix_benchmark_instance_statistics} reports the resulting benchmark-instance counts for each task and step setting. Since each toxicity cliff pair is converted into one QA instance for each of the three tasks, MolDeTox contains 52,885 instances per task and 158,655 benchmark QA instances in total. Overall, the benchmark consists of 151,749 training QA instances and 6,906 test QA instances.

Across task and step settings, multi-step instances substantially outnumber single-step instances. For Task~1, MolDeTox contains 12,204 single-step and 38,379 multi-step training instances, together with 1,091 single-step and 1,211 multi-step test instances. For Task~2 and Task~3, the benchmark contains 9,634 single-step and 40,949 multi-step training instances, together with 1,003 single-step and 1,299 multi-step test instances for each task. This distribution reflects the prevalence of compositional fragment edits in the curated toxicity cliff pairs and supports the use of MolDeTox as a benchmark not only for localized one-step detoxification, but also for more challenging multi-fragment reasoning.

Table~\ref{tab:appendix_dataset_statistics_by_source_qa} further reports source-level benchmark statistics. Among the 158,655 QA instances, Tox21 contributes the largest number of instances with 91,110 QA examples, followed by Metabolism with 25,506, AMES with 19,332, hERG with 16,065, and SIDER with 5,859. Smaller sources such as ClinTox, DICTrank, DILIst, DIRIL, and Skin Reaction are retained to preserve endpoint diversity, even though they contribute fewer QA instances.

\begin{table*}[!t]
\centering
\footnotesize
\setlength{\tabcolsep}{5pt}
\renewcommand{\arraystretch}{1.14}
\caption{
Benchmark instance statistics of MolDeTox across tasks and step settings.
}
\label{tab:appendix_benchmark_instance_statistics}
\begin{tabular}{lrr|rr|r}
\toprule
\multirow{2}{*}{\textbf{Task}}
& \multicolumn{2}{c|}{\textbf{Train}}
& \multicolumn{2}{c|}{\textbf{Test}}
& \multicolumn{1}{c}{\multirow{2}{*}{\textbf{Total}}} \\
\cmidrule(lr){2-3} \cmidrule(lr){4-5}
& \textbf{Single} & \textbf{Multi}
& \textbf{Single} & \textbf{Multi}
& \\
\midrule
Task 1: Toxic Fragment Identification & 12,204 & 38,379 & 1,091 & 1,211 & 52,885 \\
Task 2: Non-Toxic Fragment Generation  & 9,634  & 40,949 & 1,003 & 1,299 & 52,885 \\
Task 3: Non-Toxic Molecule Generation  & 9,634  & 40,949 & 1,003 & 1,299 & 52,885 \\
\midrule
\textbf{Total benchmark instances}    & 31,472 & 120,277 & 3,097 & 3,809 & 158,655 \\
\bottomrule
\end{tabular}
\end{table*}

\begin{table*}[!t]
\centering
\footnotesize
\setlength{\tabcolsep}{6pt}
\renewcommand{\arraystretch}{1.12}
\caption{
Source-level benchmark instance statistics of MolDeTox.
We report endpoint counts, molecule counts, and the number of benchmark QA instances in the train and test splits.
Raw \emph{ToxicityCliff} pair statistics are reported in Table~\ref{tab:dataset_statistics_split} of the main paper.
}
\label{tab:appendix_dataset_statistics_by_source_qa}
\begin{threeparttable}
\begin{tabular}{lrrrrr}
\toprule
\textbf{Source} & \textbf{Endpoint N} & \textbf{Molecule N} & \textbf{Train QA N} & \textbf{Test QA N} & \textbf{Total QA N} \\
\midrule
AMES          & 1  & 1,133 & 18,975 & 357   & 19,332 \\
ClinTox       & 1  & 30    & 42     & 3     & 45 \\
DICTrank      & 1  & 27    & 54     & 6     & 60 \\
DILIst        & 1  & 97    & 252    & 27    & 279 \\
DIRIL         & 1  & 6     & 0      & 9     & 9 \\
hERG          & 1  & 4,303 & 14,493 & 1,572 & 16,065 \\
Metabolism    & 5  & 3,355 & 23,421 & 2,085 & 25,506 \\
SIDER         & 25 & 212   & 4,302  & 1,557 & 5,859 \\
Skin Reaction & 1  & 56    & 372    & 18    & 390 \\
Tox21         & 12 & 1,635 & 89,838 & 1,272 & 91,110 \\
\midrule
\textbf{Total} & \textbf{49} & \textbf{10,854} & \textbf{151,749} & \textbf{6,906} & \textbf{158,655} \\
\bottomrule
\end{tabular}
\end{threeparttable}
\end{table*}
\section{Evaluation Metric Details}
\label{appendix:evaluation_metrics}

\subsection{Task 1 Metrics (Toxic Fragment Identification)}
Task~1 compares a predicted SAFE fragment string $\hat{s}$ with a gold SAFE fragment string $s$. Since a SAFE string consists of dot-separated fragments, we tokenize it by splitting on `.' after removing empty tokens and surrounding whitespace. Let $\mathcal{T}(s)$ denote the resulting fragment token list, and let $\mathrm{ms}(\mathcal{T}(s))$ denote the corresponding fragment multiset. Following our benchmark protocol, we separately report results for the single-step subset ($|\mathcal{T}(s)| = 1$) and the multi-step subset ($|\mathcal{T}(s)| \ge 2$).

\paragraph{Single / Multi Acc. (\%)}
The code computes fragment-level exact match by checking whether the predicted and gold fragment \emph{multisets} are identical:
\[
\mathrm{EM}_{\mathrm{frag}}(\hat{s}, s)=
\begin{cases}
1 & \text{if } \mathrm{ms}(\mathcal{T}(\hat{s})) = \mathrm{ms}(\mathcal{T}(s)),\\
0 & \text{otherwise}
\end{cases}
\]
The reported Acc.(\%) is obtained by averaging this exact-match indicator over the relevant subset and converting it to a percentage:
\[
\mathrm{Acc}_{\mathrm{frag}}(\%) = 100 \cdot \mathbb{E}[\mathrm{EM}_{\mathrm{frag}}]
\]

\paragraph{Multi F1}
For the Multi-step subset, we additionally compute fragment-overlap Precision, Recall, and F1 based on fragment sets. Let
\[
G = \mathrm{set}(\mathcal{T}(s)), \qquad P = \mathrm{set}(\mathcal{T}(\hat{s}))
\]
Then
\[
\mathrm{Precision} = \frac{|P \cap G|}{|P|}, \qquad
\mathrm{Recall} = \frac{|P \cap G|}{|G|},
\]
\[
\mathrm{F1} = \frac{2 \cdot \mathrm{Precision} \cdot \mathrm{Recall}}{\mathrm{Precision} + \mathrm{Recall}}
\]
We report the average F1 over the Multi-step subset.

\subsection{Task 2 Metrics (Non-Toxic Fragment Generation)}
Task~2 also compares a predicted SAFE fragment string $\hat{s}$ with a gold SAFE fragment string $s$, and likewise reports results separately for \textbf{single-step} and \textbf{multi-step} subsets.

\paragraph{Single / Multi Acc. (\%)}
Task~2 uses the same fragment-level multiset exact match as Task~1:
\[
\mathrm{EM}_{\mathrm{frag}}(\hat{s}, s)=
\begin{cases}
1 & \text{if } \mathrm{ms}(\mathcal{T}(\hat{s})) = \mathrm{ms}(\mathcal{T}(s)),\\
0 & \text{otherwise}
\end{cases}
\]
and reports
\[
\mathrm{Acc}_{\mathrm{frag}}(\%) = 100 \cdot \mathbb{E}[\mathrm{EM}_{\mathrm{frag}}]
\]
for the single-step and multi-step subsets, respectively.

\paragraph{Single / Multi Lev. Dist.}
In addition to exact match, Task~2 reports a fragment-level Levenshtein distance. This is \emph{not} computed as a single edit distance over the whole SAFE string. Instead, for each predicted fragment token $p_j \in \mathcal{T}(\hat{s})$, the code finds the minimum Levenshtein distance to any gold fragment token $g_i \in \mathcal{T}(s)$:
\[
d(p_j, s) = \min_i \mathrm{Lev}(p_j, g_i)
\]
The Task~2 fragment-level Levenshtein distance is then defined as the mean of these minimum distances over predicted fragments:
\[
\mathrm{LevFrag}(\hat{s}, s) =
\frac{1}{|\mathcal{T}(\hat{s})|}\sum_j d(p_j, s)
\]
Single Lev. Dist. and Multi Lev. Dist. are reported by averaging this value over the corresponding subsets. Lower values indicate better fragment-level replacement quality.

\paragraph{Multi F1}
For the Multi-step subset, Task~2 also reports fragment-overlap F1 using the same set-based definition as in Task~1.

\subsection{Task 3 Metrics (Non-Toxic Molecule Generation)}
Task~3 compares a predicted molecule string $\hat{m}$ with the gold non-toxic target molecule string $m_{nt}$. The code first attempts RDKit parsing and canonicalization, and then evaluates exact-match accuracy, string similarity, structural similarity, chemical validity, and a property-based score.

\paragraph{Acc. (\%)}
The code computes molecule-level exact match by checking whether the canonical SMILES strings of the predicted molecule and the gold non-toxic molecule are identical:
\[
\mathrm{EM}_{\mathrm{mol}}(\hat{m}, m_{nt})=
\begin{cases}
1 & \text{if } \mathrm{canon}(\hat{m}) = \mathrm{canon}(m_{nt}),\\
0 & \text{otherwise}
\end{cases}
\]
The reported Acc.(\%) is obtained by averaging this exact-match indicator over the relevant subset and converting it to a percentage:
\[
\mathrm{Acc}_{\mathrm{mol}}(\%) = 100 \cdot \mathbb{E}[\mathrm{EM}_{\mathrm{mol}}]
\]
Thus, Task~3 accuracy is the percentage version of exact-match correctness after canonical SMILES normalization.

\paragraph{BLEU1}
We compute unigram BLEU between the predicted molecular string and the target molecular string.

\paragraph{Levenshtein}
We compute the character-level Levenshtein distance between the predicted and target molecular strings:
\[
\mathrm{Lev}(\hat{m}, m_{nt})
\]
Lower values indicate that fewer edits are needed to transform the prediction into the target.

\paragraph{Fingerprint Similarity (RDK FTS / MACCS FTS / Morgan FTS).}
To evaluate structural similarity beyond string overlap, we compute Tanimoto similarity using three molecular fingerprints derived from canonical SMILES:
\begin{itemize}
    \item \textbf{RDK FTS:} RDKit topological fingerprint similarity
    \item \textbf{MACCS FTS:} MACCS key fingerprint similarity
    \item \textbf{Morgan FTS:} Morgan fingerprint similarity
\end{itemize}

\paragraph{Validity}
Validity is the fraction of predictions that can be parsed as chemically valid molecules by RDKit. 
For direct SMILES generation, the model output is evaluated directly as a SMILES string. 
For SAFE generation, the model output is first converted into SMILES using the deterministic SAFE-to-SMILES conversion function provided by the original SAFE implementation, and the resulting SMILES string is then parsed by RDKit. 
If SAFE-to-SMILES conversion fails, the prediction is counted as invalid and assigned a validity score of 0. 
Thus, the SAFE-to-SMILES conversion step does not mask invalid generations or introduce an additional learned decoding model; it only reconstructs SMILES from generated SAFE strings, with failed reconstructions explicitly penalized.

\paragraph{PRS}
PRS is a property-retention score that measures how well a generated non-toxic molecule preserves the drug-relevant physicochemical profile of the original toxic molecule. It is computed from six molecular properties: molecular weight (MW), logP, hydrogen-bond acceptors (HBA), hydrogen-bond donors (HBD), polar surface area (PSA), and rotatable bonds (RotB). These properties cover the main factors used in \textit{Lipinski's Rule} \cite{lipinski1997experimental} and \textit{Veber's Rule} \cite{veber2002molecular}.

For each molecule $m$, we compute a QED-inspired aggregated property score:
\[
S(m) = \sum_{i} w_i\, d_i(m), \quad i \in \{\mathrm{MW, logP, HBA, HBD, PSA, RotB}\}
\]
where $d_i(m)$ denotes the desirability function for the $i$-th physicochemical property, and $w_i$ is its corresponding weight.
For a toxic molecule $m_t$ and a generated molecule $\hat{m}$, we then measure the absolute difference between their aggregated property scores:
\[
x = \left| S(m_t) - S(\hat{m}) \right|
\]
This difference is converted into the final PRS using exponential decay:
\[
\mathrm{PRS}(\hat{m}, m_t) = \exp(-x)
\]
The reported PRS is the average over the evaluation set.

\section{Main Results with Standard Deviations}
\label{appendix:result_w_std}

\begin{table*}[h]
\centering
\scriptsize
\setlength{\tabcolsep}{3.8pt}
\renewcommand{\arraystretch}{0.95}
\caption{
Comparison of inference strategies on \textbf{Task 1} and \textbf{Task 2} of \textbf{MolDeTox} (mean $\pm$ std over three runs).
Best results are in \textbf{bold}, and second-best results are \underline{underlined}.
}
\label{tab:task12_results_std}

\resizebox{\textwidth}{!}{%
\begin{tabular}{lcccccccc}
\toprule
& \multicolumn{3}{c}{\textbf{Task 1: Toxic Frag. ID}}
& \multicolumn{5}{c}{\textbf{Task 2: NonToxic Frag. Gen.}} \\
\cmidrule(lr){2-4} \cmidrule(lr){5-9}
\textbf{Model}
& \textbf{Single Acc.(\%)}
& \textbf{Multi Acc.(\%)} & \textbf{Multi F1}
& \textbf{Single Acc.(\%)} & \textbf{Single Lev. Dist}
& \textbf{Multi Acc.(\%)} & \textbf{Multi F1} & \textbf{Multi Lev. Dist} \\
\midrule

\multicolumn{9}{l}{\textit{\textbf{LLMs}}} \\
GPT-4o
& 37.98 $\pm$ 1.30
& 1.13 $\pm$ 0.13
& 0.3789 $\pm$ 0.0074
& 5.47 $\pm$ 0.72
& 4.82 $\pm$ 0.33
& 0.18 $\pm$ 0.12
& 0.0141 $\pm$ 0.0033
& 4.87 $\pm$ 0.26 \\
GPT-5.2
& 41.09 $\pm$ 0.28
& 3.85 $\pm$ 0.10
& 0.4358 $\pm$ 0.0008
& \underline{7.92 $\pm$ 0.15}
& \underline{3.96 $\pm$ 0.07}
& 0.23 $\pm$ 0.08
& 0.0236 $\pm$ 0.0014
& \underline{4.40 $\pm$ 0.02} \\
GPT-5.2 4-Shot
& \textbf{54.29 $\pm$ 0.61}
& \textbf{14.49 $\pm$ 0.54}
& \underline{0.5562 $\pm$ 0.0016}
& \textbf{21.36 $\pm$ 0.49}
& \textbf{3.62 $\pm$ 0.03}
& \textbf{4.77 $\pm$ 0.14}
& \textbf{0.1215 $\pm$ 0.0012}
& \textbf{4.19 $\pm$ 0.06} \\
Qwen3-4B-Inst.
& 40.56 $\pm$ 8.48
& 0.94 $\pm$ 1.62
& 0.4732 $\pm$ 0.0319
& 0.33 $\pm$ 0.57
& 4.96 $\pm$ 0.09
& 0.00 $\pm$ 0.00
& 0.0005 $\pm$ 0.0008
& 5.78 $\pm$ 0.04 \\
Qwen3-4B-Inst. 4-Shot
& 36.40 $\pm$ 0.13
& \underline{7.57 $\pm$ 0.19}
& 0.5122 $\pm$ 0.0012
& 7.80 $\pm$ 0.24
& 4.65 $\pm$ 0.01
& \underline{0.63 $\pm$ 0.05}
& \underline{0.0305 $\pm$ 0.0011}
& 5.28 $\pm$ 0.01 \\
Qwen3-8B
& 34.87 $\pm$ 2.58
& 3.58 $\pm$ 6.20
& \textbf{0.5648 $\pm$ 0.0216}
& 0.73 $\pm$ 1.26
& 6.79 $\pm$ 4.14
& 0.05 $\pm$ 0.09
& 0.0009 $\pm$ 0.0016
& 4.92 $\pm$ 0.05 \\
Llama-3.1-8B-Inst.
& 17.51 $\pm$ 4.26
& 5.54 $\pm$ 1.99
& 0.4782 $\pm$ 0.0215
& 0.53 $\pm$ 0.25
& 28.69 $\pm$ 12.86
& 0.00 $\pm$ 0.00
& 0.0013 $\pm$ 0.0011
& 40.10 $\pm$ 7.80 \\
Gemma-3-27B
& \underline{42.45 $\pm$ 0.45}
& 5.79 $\pm$ 0.38
& 0.4888 $\pm$ 0.0021
& 2.05 $\pm$ 0.25
& 6.83 $\pm$ 0.07
& 0.00 $\pm$ 0.00
& 0.0012 $\pm$ 0.0004
& 8.88 $\pm$ 0.09 \\
DeepSeek-Llama-70B
& 37.74 $\pm$ 0.96
& 3.84 $\pm$ 0.40
& 0.4506 $\pm$ 0.0035
& 3.43 $\pm$ 0.21
& 5.10 $\pm$ 0.10
& 0.04 $\pm$ 0.05
& 0.0092 $\pm$ 0.0017
& 5.97 $\pm$ 0.14 \\

\addlinespace[2pt]
\multicolumn{9}{l}{\textit{\textbf{VLMs}}} \\
GPT-5.2 w/ Image
& \underline{40.31 $\pm$ 0.40}
& \underline{4.30 $\pm$ 0.25}
& 0.4467 $\pm$ 0.0011
& \textbf{8.15 $\pm$ 0.10}
& \textbf{3.93 $\pm$ 0.04}
& \textbf{0.18 $\pm$ 0.09}
& \textbf{0.0232 $\pm$ 0.0004}
& \textbf{4.38 $\pm$ 0.04} \\
Gemini 3.1 Flash-Lite w/ Image
& \textbf{43.78 $\pm$ 0.93}
& \textbf{4.65 $\pm$ 0.33}
& 0.4361 $\pm$ 0.0027
& \underline{6.69 $\pm$ 0.32}
& \underline{4.40 $\pm$ 0.06}
& \underline{0.08 $\pm$ 0.00}
& \underline{0.0218 $\pm$ 0.0026}
& \underline{4.73 $\pm$ 0.01} \\
Qwen3-VL-4B-Inst.
& 27.73 $\pm$ 0.16
& 3.50 $\pm$ 0.31
& \underline{0.5763 $\pm$ 0.0017}
& 0.13 $\pm$ 0.06
& 5.75 $\pm$ 0.05
& 0.00 $\pm$ 0.00
& 0.0005 $\pm$ 0.0002
& 7.08 $\pm$ 2.44 \\
Qwen3-VL-4B-Inst. w/ Color Image
& 27.78 $\pm$ 0.84
& 3.06 $\pm$ 0.14
& \textbf{0.5825 $\pm$ 0.0039}
& 0.10 $\pm$ 0.10
& 5.57 $\pm$ 0.06
& 0.00 $\pm$ 0.00
& 0.0001 $\pm$ 0.0002
& 5.59 $\pm$ 0.02 \\
Qwen3-VL-8B-Inst.
& 33.46 $\pm$ 2.52
& 2.01 $\pm$ 1.75
& 0.5005 $\pm$ 0.0243
& 0.80 $\pm$ 0.69
& 4.50 $\pm$ 1.67
& 0.00 $\pm$ 0.00
& 0.0005 $\pm$ 0.0006
& 6.05 $\pm$ 0.05 \\
LLaVA-v1.6-Vicuna-13B
& 3.31 $\pm$ 2.90
& 0.30 $\pm$ 0.33
& 0.4013 $\pm$ 0.0456
& 0.00 $\pm$ 0.00
& 12.69 $\pm$ 4.29
& 0.00 $\pm$ 0.00
& 0.0014 $\pm$ 0.0012
& 9.19 $\pm$ 5.54 \\

\bottomrule
\end{tabular}%
}
\end{table*}

\begin{table*}[h]
\centering
\scriptsize
\setlength{\tabcolsep}{4.0pt}
\renewcommand{\arraystretch}{0.95}
\caption{
Comparison of inference strategies on \textbf{Task 3} of \textbf{MolDeTox} (mean $\pm$ std over three runs).
We compare model performance under single-step and multi-step settings across SMILES and SAFE generation for both LLMs and VLMs.
Best results are in \textbf{bold}, and second-best results are \underline{underlined}.
}
\label{tab:task3_std_merged}

\resizebox{\textwidth}{!}{%
\begin{tabular}{lcccccccc}
\toprule
\textbf{Model} & \textbf{Acc.(\%)} & \textbf{BLEU1} & \textbf{Levenshtein} & \textbf{RDK FTS} & \textbf{MACCS FTS} & \textbf{Morgan FTS} & \textbf{Validity} & \textbf{PRS} \\
\midrule

\rowcolor{gray!12}
\multicolumn{9}{c}{\textbf{Single Step}} \\

\multicolumn{9}{l}{\textit{\textbf{SMILES Generation}}} \\
\multicolumn{9}{l}{\textit{\textbf{LLMs}}} \\
GPT-5.2
& \textbf{3.50 $\pm$ 0.30}
& \underline{0.941 $\pm$ 0.002}
& \textbf{7.29 $\pm$ 0.14}
& 0.702 $\pm$ 0.003
& \underline{0.753 $\pm$ 0.003}
& 0.557 $\pm$ 0.003
& \textbf{0.930 $\pm$ 0.001}
& \underline{0.770 $\pm$ 0.061} \\
Qwen3-4B-Inst.
& 0.13 $\pm$ 0.23
& 0.810 $\pm$ 0.021
& \underline{22.03 $\pm$ 3.25}
& \textbf{0.794 $\pm$ 0.040}
& \textbf{0.796 $\pm$ 0.010}
& \textbf{0.648 $\pm$ 0.012}
& \underline{0.918 $\pm$ 0.015}
& \textbf{0.775 $\pm$ 0.026} \\
Qwen3-8B
& 0.40 $\pm$ 0.69
& \textbf{0.959 $\pm$ 0.015}
& 35.07 $\pm$ 41.86
& \underline{0.730 $\pm$ 0.031}
& 0.709 $\pm$ 0.024
& \underline{0.592 $\pm$ 0.010}
& 0.848 $\pm$ 0.051
& 0.655 $\pm$ 0.047 \\
DeepSeek-Llama-70B
& \underline{1.88 $\pm$ 0.00}
& 0.898 $\pm$ 0.000
& 556.68 $\pm$ 0.00
& 0.463 $\pm$ 0.000
& 0.495 $\pm$ 0.000
& 0.358 $\pm$ 0.000
& 0.703 $\pm$ 0.000
& 0.548 $\pm$ 0.000 \\

\addlinespace[2pt]
\multicolumn{9}{l}{\textit{\textbf{SAFE Generation}}} \\
\multicolumn{9}{l}{\textit{\textbf{LLMs}}} \\
GPT-4o
& 1.98 $\pm$ 0.45
& 0.853 $\pm$ 0.011
& 8.22 $\pm$ 0.23
& 0.670 $\pm$ 0.009
& 0.724 $\pm$ 0.011
& 0.536 $\pm$ 0.008
& 0.915 $\pm$ 0.011
& 0.725 $\pm$ 0.010 \\
GPT-5.2
& \textbf{4.00 $\pm$ 0.11}
& 0.883 $\pm$ 0.006
& 6.70 $\pm$ 0.14
& 0.720 $\pm$ 0.006
& 0.762 $\pm$ 0.006
& 0.583 $\pm$ 0.006
& 0.936 $\pm$ 0.006
& 0.741 $\pm$ 0.006 \\
GPT-5.2 w/ CoT
& 3.77 $\pm$ 0.36
& 0.885 $\pm$ 0.006
& 6.63 $\pm$ 0.13
& 0.711 $\pm$ 0.007
& 0.754 $\pm$ 0.006
& 0.577 $\pm$ 0.006
& 0.940 $\pm$ 0.006
& 0.743 $\pm$ 0.004 \\
GPT-5.2 4-Shot
& \textbf{15.59 $\pm$ 0.40}
& 0.914 $\pm$ 0.006
& 6.44 $\pm$ 0.22
& 0.777 $\pm$ 0.003
& 0.816 $\pm$ 0.006
& 0.652 $\pm$ 0.002
& 0.969 $\pm$ 0.007
& \underline{0.922 $\pm$ 0.008} \\
Gemini 3.1 Flash-Lite
& 0.43 $\pm$ 0.35
& 0.605 $\pm$ 0.026
& 29.94 $\pm$ 2.08
& 0.356 $\pm$ 0.036
& 0.346 $\pm$ 0.043
& 0.154 $\pm$ 0.041
& 0.887 $\pm$ 0.005
& 0.691 $\pm$ 0.003 \\
Qwen3-4B-Inst.
& 0.07 $\pm$ 0.11
& \textbf{0.954 $\pm$ 0.010}
& \underline{3.50 $\pm$ 1.02}
& \textbf{0.860 $\pm$ 0.040}
& \textbf{0.868 $\pm$ 0.006}
& \textbf{0.709 $\pm$ 0.016}
& \textbf{0.998 $\pm$ 0.003}
& 0.817 $\pm$ 0.026 \\
Qwen3-4B-Inst. w/ CoT
& 0.43 $\pm$ 0.74
& 0.784 $\pm$ 0.006
& 9.75 $\pm$ 1.11
& 0.644 $\pm$ 0.050
& 0.687 $\pm$ 0.011
& 0.494 $\pm$ 0.020
& 0.862 $\pm$ 0.046
& 0.675 $\pm$ 0.035 \\
Qwen3-4B-Inst. 4-Shot
& 3.38 $\pm$ 0.05
& \underline{0.942 $\pm$ 0.002}
& 5.18 $\pm$ 0.02
& \underline{0.822 $\pm$ 0.001}
& \underline{0.865 $\pm$ 0.001}
& \underline{0.699 $\pm$ 0.001}
& \underline{0.994 $\pm$ 0.001}
& \textbf{0.982 $\pm$ 0.002} \\
Qwen3-8B
& 0.00 $\pm$ 0.00
& 0.643 $\pm$ 0.548
& \textbf{1.95 $\pm$ 1.66}
& 0.592 $\pm$ 0.504
& 0.584 $\pm$ 0.497
& 0.482 $\pm$ 0.410
& 0.670 $\pm$ 0.571
& \textbf{0.832 $\pm$ 0.000} \\
Llama-3.1-8B-Inst.
& 0.00 $\pm$ 0.00
& 0.290 $\pm$ 0.247
& 19.00 $\pm$ 16.14
& 0.091 $\pm$ 0.077
& 0.090 $\pm$ 0.076
& 0.051 $\pm$ 0.043
& 0.532 $\pm$ 0.453
& 0.585 $\pm$ 0.049 \\
Gemma-3-27B
& 1.55 $\pm$ 0.06
& 0.853 $\pm$ 0.003
& 7.39 $\pm$ 0.27
& 0.731 $\pm$ 0.003
& 0.779 $\pm$ 0.004
& 0.596 $\pm$ 0.002
& 0.942 $\pm$ 0.004
& 0.733 $\pm$ 0.003 \\
DeepSeek-Llama-70B
& 2.30 $\pm$ 0.62
& 0.712 $\pm$ 0.013
& 11.10 $\pm$ 0.07
& 0.519 $\pm$ 0.010
& 0.555 $\pm$ 0.010
& 0.392 $\pm$ 0.007
& 0.780 $\pm$ 0.017
& 0.614 $\pm$ 0.010 \\

\multicolumn{9}{l}{\textit{\textbf{VLMs}}} \\
GPT-5.2 w/ Image
& \underline{2.61 $\pm$ 2.27}
& 0.894 $\pm$ 0.047
& 7.49 $\pm$ 1.23
& 0.733 $\pm$ 0.055
& 0.766 $\pm$ 0.036
& 0.587 $\pm$ 0.036
& 0.945 $\pm$ 0.048
& 0.765 $\pm$ 0.070 \\
Gemini 3.1 Flash-Lite w/ Image
& \textbf{3.19 $\pm$ 0.35}
& 0.823 $\pm$ 0.001
& 9.68 $\pm$ 0.06
& 0.660 $\pm$ 0.002
& 0.702 $\pm$ 0.002
& 0.508 $\pm$ 0.004
& 0.888 $\pm$ 0.002
& 0.693 $\pm$ 0.002 \\
Qwen3-VL-4B-Inst.
& 0.07 $\pm$ 0.06
& \underline{0.929 $\pm$ 0.001}
& \underline{4.74 $\pm$ 0.03}
& \underline{0.789 $\pm$ 0.001}
& \underline{0.843 $\pm$ 0.001}
& \underline{0.671 $\pm$ 0.000}
& \underline{0.983 $\pm$ 0.002}
& \underline{0.778 $\pm$ 0.001} \\
Qwen3-VL-4B-Inst. w/ Color Image
& 0.10 $\pm$ 0.10
& \textbf{0.936 $\pm$ 0.020}
& \textbf{4.08 $\pm$ 1.02}
& \textbf{0.820 $\pm$ 0.055}
& \textbf{0.849 $\pm$ 0.019}
& \textbf{0.685 $\pm$ 0.029}
& \textbf{0.987 $\pm$ 0.012}
& \textbf{0.793 $\pm$ 0.034} \\
Qwen3-VL-8B-Inst.
& 0.50 $\pm$ 0.45
& 0.908 $\pm$ 0.043
& 6.53 $\pm$ 2.67
& 0.776 $\pm$ 0.082
& 0.813 $\pm$ 0.056
& 0.637 $\pm$ 0.061
& 0.966 $\pm$ 0.029
& 0.776 $\pm$ 0.050 \\
LLaVA-v1.6-Vicuna-13B
& 0.00 $\pm$ 0.00
& 0.356 $\pm$ 0.309
& 18.58 $\pm$ 15.81
& 0.121 $\pm$ 0.104
& 0.154 $\pm$ 0.140
& 0.081 $\pm$ 0.070
& 0.538 $\pm$ 0.459
& 0.391 $\pm$ 0.335 \\

\midrule
\rowcolor{gray!12}
\multicolumn{9}{c}{\textbf{Multi Step}} \\

\multicolumn{9}{l}{\textit{\textbf{SMILES Generation}}} \\
\multicolumn{9}{l}{\textit{\textbf{LLMs}}} \\
GPT-5.2
& 0.21 $\pm$ 0.16
& \underline{0.894 $\pm$ 0.001}
& \underline{13.17 $\pm$ 0.02}
& 0.588 $\pm$ 0.005
& \textbf{0.628 $\pm$ 0.003}
& 0.408 $\pm$ 0.003
& \textbf{0.933 $\pm$ 0.007}
& \textbf{0.721 $\pm$ 0.005} \\
Qwen3-4B-Inst.
& 0.00 $\pm$ 0.00
& 0.862 $\pm$ 0.060
& 15.30 $\pm$ 4.43
& \textbf{0.640 $\pm$ 0.066}
& \underline{0.625 $\pm$ 0.052}
& \textbf{0.445 $\pm$ 0.025}
& 0.843 $\pm$ 0.017
& 0.592 $\pm$ 0.104 \\
Qwen3-8B
& \underline{0.28 $\pm$ 0.49}
& \textbf{0.941 $\pm$ 0.037}
& \textbf{11.99 $\pm$ 2.87}
& \underline{0.618 $\pm$ 0.043}
& 0.617 $\pm$ 0.035
& \underline{0.418 $\pm$ 0.027}
& \underline{0.856 $\pm$ 0.039}
& \underline{0.602 $\pm$ 0.071} \\
DeepSeek-Llama-70B
& \textbf{0.60 $\pm$ 0.00}
& 0.862 $\pm$ 0.000
& 530.28 $\pm$ 0.00
& 0.394 $\pm$ 0.000
& 0.414 $\pm$ 0.000
& 0.265 $\pm$ 0.000
& 0.682 $\pm$ 0.000
& 0.519 $\pm$ 0.000 \\

\addlinespace[2pt]
\multicolumn{9}{l}{\textit{\textbf{SAFE Generation}}} \\
\multicolumn{9}{l}{\textit{\textbf{LLMs}}} \\
GPT-4o
& 0.10 $\pm$ 0.04
& 0.814 $\pm$ 0.005
& 13.11 $\pm$ 0.18
& 0.564 $\pm$ 0.003
& 0.603 $\pm$ 0.004
& 0.385 $\pm$ 0.005
& 0.922 $\pm$ 0.004
& 0.714 $\pm$ 0.004 \\
GPT-5.2
& 0.15 $\pm$ 0.15
& 0.855 $\pm$ 0.004
& 11.28 $\pm$ 0.08
& 0.611 $\pm$ 0.002
& 0.650 $\pm$ 0.002
& 0.435 $\pm$ 0.003
& 0.953 $\pm$ 0.003
& 0.738 $\pm$ 0.002 \\
GPT-5.2 w/ CoT
& 0.23 $\pm$ 0.08
& 0.842 $\pm$ 0.001
& 11.60 $\pm$ 0.14
& 0.593 $\pm$ 0.001
& 0.635 $\pm$ 0.002
& 0.419 $\pm$ 0.001
& 0.942 $\pm$ 0.002
& 0.729 $\pm$ 0.003 \\
GPT-5.2 4-Shot
& \textbf{2.95 $\pm$ 0.21}
& 0.860 $\pm$ 0.001
& 11.60 $\pm$ 0.02
& 0.647 $\pm$ 0.002
& 0.683 $\pm$ 0.001
& 0.478 $\pm$ 0.002
& 0.956 $\pm$ 0.001
& \underline{0.899 $\pm$ 0.001} \\
Gemini 3.1 Flash-Lite
& 0.21 $\pm$ 0.16
& 0.718 $\pm$ 0.031
& 23.27 $\pm$ 3.01
& 0.439 $\pm$ 0.045
& 0.433 $\pm$ 0.060
& 0.235 $\pm$ 0.051
& 0.935 $\pm$ 0.005
& 0.741 $\pm$ 0.004 \\
Qwen3-4B-Inst.
& 0.00 $\pm$ 0.00
& \textbf{0.941 $\pm$ 0.051}
& \underline{10.27 $\pm$ 0.76}
& \textbf{0.755 $\pm$ 0.084}
& \textbf{0.787 $\pm$ 0.075}
& \textbf{0.528 $\pm$ 0.040}
& \underline{0.997 $\pm$ 0.004}
& 0.744 $\pm$ 0.023 \\
Qwen3-4B-Inst. w/ CoT
& 0.03 $\pm$ 0.04
& 0.823 $\pm$ 0.075
& 13.16 $\pm$ 2.26
& 0.591 $\pm$ 0.055
& 0.644 $\pm$ 0.084
& 0.402 $\pm$ 0.034
& 0.909 $\pm$ 0.084
& 0.656 $\pm$ 0.058 \\
Qwen3-4B-Inst. 4-Shot
& \underline{0.49 $\pm$ 0.00}
& \underline{0.897 $\pm$ 0.000}
& 10.92 $\pm$ 0.03
& \underline{0.692 $\pm$ 0.001}
& \underline{0.732 $\pm$ 0.000}
& \underline{0.519 $\pm$ 0.001}
& \textbf{0.999 $\pm$ 0.000}
& \textbf{0.985 $\pm$ 0.000} \\
Qwen3-8B
& 0.00 $\pm$ 0.00
& 0.625 $\pm$ 0.538
& \textbf{7.85 $\pm$ 7.07}
& 0.493 $\pm$ 0.425
& 0.504 $\pm$ 0.433
& 0.348 $\pm$ 0.300
& 0.668 $\pm$ 0.575
& 0.734 $\pm$ 0.008 \\
Llama-3.1-8B-Inst.
& 0.00 $\pm$ 0.00
& 0.289 $\pm$ 0.250
& 17.95 $\pm$ 15.53
& 0.090 $\pm$ 0.078
& 0.095 $\pm$ 0.082
& 0.053 $\pm$ 0.046
& 0.546 $\pm$ 0.472
& 0.447 $\pm$ 0.190 \\
Gemma-3-27B
& 0.05 $\pm$ 0.04
& 0.797 $\pm$ 0.003
& 13.24 $\pm$ 0.18
& 0.595 $\pm$ 0.002
& 0.635 $\pm$ 0.002
& 0.417 $\pm$ 0.000
& 0.933 $\pm$ 0.002
& 0.711 $\pm$ 0.003 \\
DeepSeek-Llama-70B
& 0.26 $\pm$ 0.05
& 0.696 $\pm$ 0.007
& 12.85 $\pm$ 0.14
& 0.459 $\pm$ 0.004
& 0.478 $\pm$ 0.002
& 0.299 $\pm$ 0.005
& 0.789 $\pm$ 0.005
& 0.610 $\pm$ 0.005 \\

\multicolumn{9}{l}{\textit{\textbf{VLMs}}} \\
GPT-5.2 w/ Image
& 0.15 $\pm$ 0.20
& \underline{0.889 $\pm$ 0.078}
& \underline{10.53 $\pm$ 1.18}
& \underline{0.648 $\pm$ 0.095}
& \underline{0.686 $\pm$ 0.085}
& 0.445 $\pm$ 0.040
& 0.959 $\pm$ 0.036
& 0.722 $\pm$ 0.010 \\
Gemini 3.1 Flash-Lite w/ Image
& \underline{0.21 $\pm$ 0.04}
& 0.798 $\pm$ 0.009
& 13.89 $\pm$ 0.03
& 0.571 $\pm$ 0.007
& 0.604 $\pm$ 0.004
& 0.384 $\pm$ 0.004
& 0.914 $\pm$ 0.009
& 0.727 $\pm$ 0.008 \\
Qwen3-VL-4B-Inst.
& 0.03 $\pm$ 0.04
& 0.868 $\pm$ 0.005
& 11.67 $\pm$ 0.36
& 0.636 $\pm$ 0.015
& 0.678 $\pm$ 0.018
& \underline{0.464 $\pm$ 0.016}
& \underline{0.984 $\pm$ 0.001}
& \textbf{0.755 $\pm$ 0.009} \\
Qwen3-VL-4B-Inst. w/ Color Image
& 0.00 $\pm$ 0.00
& \textbf{0.894 $\pm$ 0.042}
& \textbf{10.49 $\pm$ 1.87}
& \textbf{0.691 $\pm$ 0.091}
& \textbf{0.725 $\pm$ 0.076}
& \textbf{0.509 $\pm$ 0.072}
& \textbf{0.989 $\pm$ 0.010}
& \underline{0.750 $\pm$ 0.012} \\
Qwen3-VL-8B-Inst.
& \textbf{0.41 $\pm$ 0.36}
& 0.790 $\pm$ 0.062
& 12.45 $\pm$ 1.84
& 0.556 $\pm$ 0.043
& 0.583 $\pm$ 0.070
& 0.393 $\pm$ 0.039
& 0.907 $\pm$ 0.064
& 0.714 $\pm$ 0.023 \\
LLaVA-v1.6-Vicuna-13B
& 0.00 $\pm$ 0.00
& 0.298 $\pm$ 0.256
& 13.88 $\pm$ 12.97
& 0.127 $\pm$ 0.117
& 0.151 $\pm$ 0.146
& 0.085 $\pm$ 0.080
& 0.487 $\pm$ 0.422
& 0.334 $\pm$ 0.289 \\

\bottomrule
\end{tabular}%
}
\end{table*}
\clearpage

\newpage
\section{Endpoint-wise Result Analysis}
\label{appendix:endpoint_wise_result_analysis}

\begin{table*}[h]
\centering
\scriptsize
\setlength{\tabcolsep}{4.0pt}
\renewcommand{\arraystretch}{0.96}
\caption{
Main results on \textbf{MolDeTox}, grouped by endpoint.
The upper block reports results for \textbf{Task 1} and \textbf{Task 2}, and the lower block reports \textbf{Task 3} (\textbf{SAFE Generation}).
Higher is better for Acc.(\%) and F1, while lower is better for Levenshtein Dist.
For Task 3, higher is better for Acc.(\%), Morgan FTS, Validity, and PRS, while lower is better for Levenshtein.
}
\label{tab:results_by_endpoint}

\begin{tabularx}{\textwidth}{l l Y|YYYY|YYY}
\toprule
& & \multicolumn{3}{c}{\textbf{Task 1: Toxic Frag. ID}}
& \multicolumn{5}{c}{\textbf{Task 2: NonToxic Frag. Gen.}} \\
\cmidrule(lr){3-5} \cmidrule(lr){6-10}
\textbf{Model} & \textbf{Endpoint}
& \multicolumn{1}{c|}{\textbf{Single}}
& \multicolumn{2}{c}{\textbf{Multi}}
& \multicolumn{2}{c|}{\textbf{Single}}
& \multicolumn{3}{c}{\textbf{Multi}} \\
&
& \textbf{Acc.(\%)}
& \textbf{Acc.(\%)} & \textbf{F1}
& \textbf{Acc.(\%)} & \textbf{Lev. Dist}
& \textbf{Acc.(\%)} & \textbf{F1} & \textbf{Lev. Dist} \\
\midrule

\multirow{6}{*}{GPT-5.2}
& herg\_unified   & 36.15 & 3.07  & 0.3547 & 5.92 & 3.99 & \underline{0.42} & \textbf{0.0390} & \underline{4.16} \\
& cyp2c19\_veith & 32.53 & 0.00  & 0.3717 & \underline{6.49} & 3.78 & 0.00 & \underline{0.0309} & 4.25 \\
& cyp1a2\_veith  & 44.93 & 5.13  & 0.4443 & \textbf{8.62} & \textbf{3.43} & 0.00 & 0.0225 & \textbf{4.06} \\
& ames           & \underline{69.35} & 8.77  & \underline{0.5644} & 5.26 & \underline{3.65} & \textbf{1.61} & 0.0242 & 4.56 \\
& tox21\_NR-ER   & 36.36 & \textbf{14.29} & \textbf{0.6223} & 5.26 & 4.37 & 0.00 & 0.0000 & 4.35 \\
& tox21\_SR-MMP  & \textbf{70.00} & \underline{10.34} & 0.4461 & 0.00 & 6.06 & 0.00 & 0.0000 & 4.41 \\
\midrule

\multirow{6}{*}{Qwen3-4B-Inst.}
& herg\_unified   & \underline{28.04} & 3.51 & 0.3505 & \underline{1.05} & 4.91 & 0.00 & 0.0000 & 6.39 \\
& cyp2c19\_veith & 16.87 & 0.00 & 0.3919 & 0.00 & \underline{4.77} & 0.00 & \underline{0.0032} & \textbf{5.71} \\
& cyp1a2\_veith  & 18.84 & 1.28 & 0.3663 & 0.00 & \textbf{3.95} & 0.00 & \underline{0.0032} & \underline{5.89} \\
& ames           & \textbf{43.55} & \textbf{8.77} & 0.4500 & \textbf{1.75} & 6.91 & 0.00 & \textbf{0.0081} & \underline{5.93} \\
& tox21\_NR-ER   & 27.27 & \underline{3.57} & \textbf{0.5520} & 0.00 & 4.89 & 0.00 & 0.0000 & 6.06 \\
& tox21\_SR-MMP  & 25.00 & 3.45 & \underline{0.5103} & 0.00 & 6.62 & 0.00 & 0.0000 & \underline{5.93} \\
\midrule

\multirow{6}{*}{GPT-5.2 w/ Image}
& herg\_unified   & 31.76 & 3.95  & 0.3704 & 6.27  & 3.85 & 0.00 & \textbf{0.0323} & 4.18 \\
& cyp2c19\_veith & 30.12 & 2.41  & 0.3887 & \underline{9.09}  & 3.77 & 0.00 & 0.0253 & 4.16 \\
& cyp1a2\_veith  & 42.03 & 3.85  & 0.4625 & \textbf{13.79} & \textbf{3.29} & 0.00 & \underline{0.0285} & \underline{4.06} \\
& ames           & \textbf{72.58} & \underline{10.53} & \textbf{0.5912} & 5.26  & \underline{3.63} & 0.00 & 0.0081 & 4.50 \\
& tox21\_NR-ER   & 36.36 & \textbf{14.29} & \underline{0.5607} & 5.26  & 4.58 & 0.00 & 0.0269 & \textbf{3.87} \\
& tox21\_SR-MMP  & \underline{55.00} & 10.34 & 0.5230 & 0.00  & 5.94 & 0.00 & 0.0000 & 4.33 \\
\midrule

\multirow{6}{*}{Qwen3-VL-4B-Inst.}
& herg\_unified   & 26.01 & 3.07  & 0.4664 & 0.00 & \underline{5.09}  & 0.00 & 0.0000 & 29.41 \\
& cyp2c19\_veith & 15.66 & 6.02  & 0.5466 & 0.00 & \textbf{4.60}  & 0.00 & 0.0000 & \underline{5.42} \\
& cyp1a2\_veith  & 15.94 & 3.85  & 0.5131 & 0.00 & 5.10  & 0.00 & 0.0000 & 5.68 \\
& ames           & \textbf{50.00} & \underline{7.02}  & \textbf{0.6947} & 0.00 & 11.88 & 0.00 & \textbf{0.0081} & 5.55 \\
& tox21\_NR-ER   & \underline{40.91} & \textbf{10.71} & \underline{0.6896} & 0.00 & 7.09  & 0.00 & 0.0000 & 5.62 \\
& tox21\_SR-MMP  & 40.00 & 0.00  & 0.6268 & 0.00 & 8.69  & 0.00 & 0.0000 & \textbf{4.74} \\
\bottomrule
\end{tabularx}

\vspace{0.8em}

\begin{tabularx}{\textwidth}{l l YYYYY|YYYYY}
\toprule
& & \multicolumn{10}{c}{\textbf{Task 3: Nontoxic Molecule Generation (SAFE Generation)}} \\
\cmidrule(lr){3-12}
\textbf{Model} & \textbf{Endpoint}
& \multicolumn{5}{c|}{\textbf{Single-Step}}
& \multicolumn{5}{c}{\textbf{Multi-Step}} \\
\cmidrule(lr){3-7} \cmidrule(lr){8-12}
&
& \textbf{Acc.(\%)} & \textbf{Lev.} & \textbf{Morgan FTS} & \textbf{Val.} & \textbf{PRS}
& \textbf{Acc.(\%)} & \textbf{Lev.} & \textbf{Morgan FTS} & \textbf{Val.} & \textbf{PRS} \\
\midrule

\multirow{6}{*}{GPT-5.2}
& herg\_unified   & 3.14 & 7.93 & \textbf{0.634} & 0.930 & 0.748 & 0.00 & \underline{10.55} & \textbf{0.534} & 0.949 & 0.764 \\
& cyp2c19\_veith & 2.60 & \underline{6.23} & 0.591 & 0.922 & 0.733 & 0.00 & 11.06 & 0.446 & 0.899 & 0.724 \\
& cyp1a2\_veith  & \textbf{5.17} & 7.03 & \underline{0.619} & \textbf{0.948} & \textbf{0.780} & 0.00 & 10.87 & \underline{0.509} & \textbf{0.978} & \textbf{0.804} \\
& ames           & \underline{3.51} & \textbf{5.89} & 0.498 & \underline{0.947} & \underline{0.752} & \textbf{1.61} & \textbf{9.74}  & 0.354 & \underline{0.968} & \underline{0.777} \\
& tox21\_NR-ER   & 0.00 & 9.58 & 0.441 & 0.895 & 0.707 & 0.00 & 11.58 & 0.410 & 0.935 & \underline{0.777} \\
& tox21\_SR-MMP  & 0.00 & 8.94 & 0.363 & 0.824 & 0.645 & 0.00 & 13.19 & 0.296 & 0.906 & 0.717 \\
\midrule

\multirow{6}{*}{Qwen3-4B-Inst.}
& herg\_unified   & 0.00 & 5.12 & 0.732 & \underline{0.993} & 0.790 & 0.00 & \underline{9.77}  & \textbf{0.598} & \underline{0.992} & 0.802 \\
& cyp2c19\_veith & 0.00 & \textbf{3.52} & \textbf{0.741} & \textbf{1.000} & 0.795 & 0.00 & 11.06 & 0.534 & \textbf{1.000} & 0.807 \\
& cyp1a2\_veith  & \textbf{1.72} & 3.90 & \underline{0.737} & \textbf{1.000} & \textbf{0.820} & 0.00 & 10.08 & \underline{0.540} & \textbf{1.000} & \textbf{0.819} \\
& ames           & 0.00 & 4.32 & 0.561 & 0.982 & 0.782 & 0.00 & 10.79 & 0.398 & 0.984 & \underline{0.809} \\
& tox21\_NR-ER   & 0.00 & \underline{3.89} & 0.624 & 0.947 & 0.744 & 0.00 & \textbf{9.19}  & 0.464 & 0.968 & 0.801 \\
& tox21\_SR-MMP  & 0.00 & 6.24 & 0.560 & \textbf{1.000} & \underline{0.799} & 0.00 & 12.31 & 0.336 & 0.906 & 0.712 \\
\midrule

\multirow{6}{*}{GPT-5.2 w/ Image}
& herg\_unified   & 3.14 & 6.89 & \textbf{0.606} & 0.899 & 0.720 & 0.00 & 9.96  & \textbf{0.526} & 0.932 & 0.753 \\
& cyp2c19\_veith & 3.90 & \underline{5.78} & \underline{0.566} & 0.883 & 0.704 & 0.00 & 11.49 & 0.471 & 0.921 & 0.742 \\
& cyp1a2\_veith  & \underline{5.17} & 6.50 & 0.552 & 0.845 & 0.684 & 0.00 & 11.36 & \underline{0.477} & \underline{0.966} & \textbf{0.802} \\
& ames           & \textbf{5.26} & \textbf{5.04} & 0.524 & \underline{0.965} & 0.763 & 0.00 & \underline{9.90}  & 0.348 & \textbf{0.984} & \underline{0.786} \\
& tox21\_NR-ER   & 0.00 & 12.26 & 0.482 & \textbf{1.000} & \textbf{0.791} & \textbf{3.23} & \textbf{9.42} & 0.397 & \underline{0.935} & 0.782 \\
& tox21\_SR-MMP  & 0.00 & 10.35 & 0.465 & \textbf{1.000} & \underline{0.781} & 0.00 & 12.38 & 0.262 & 0.812 & \underline{0.781} \\
\midrule

\multirow{6}{*}{Qwen3-VL-4B-Inst.}
& herg\_unified   & 0.00 & 4.96 & \textbf{0.725} & \underline{0.990} & \underline{0.792} & 0.00 & \textbf{9.43}  & \textbf{0.595} & 0.985 & 0.795 \\
& cyp2c19\_veith & 0.00 & \textbf{3.70} & \underline{0.715} & \underline{0.974} & 0.774 & 0.00 & 13.11 & 0.518 & \underline{0.990} & 0.806 \\
& cyp1a2\_veith  & 0.00 & \underline{4.19} & 0.693 & 0.966 & \underline{0.789} & 0.00 & 10.81 & \underline{0.536} & \textbf{1.000} & \textbf{0.818} \\
& ames           & 0.00 & 6.18 & 0.516 & 0.982 & 0.778 & 0.00 & \underline{10.07} & 0.393 & 0.972 & 0.787 \\
& tox21\_NR-ER   & 0.00 & 4.47 & 0.649 & \textbf{1.000} & 0.781 & 0.00 & 12.76 & 0.429 & \textbf{1.000} & \underline{0.813} \\
& tox21\_SR-MMP  & 0.00 & 7.06 & 0.553 & \textbf{1.000} & \textbf{0.805} & 0.00 & 15.74 & 0.343 & 0.971 & 0.750 \\
\bottomrule
\end{tabularx}
\end{table*}

Table~\ref{tab:results_by_endpoint} reports endpoint-wise results for Task~1, Task~2, and Task~3. Overall, model performance varies substantially across endpoints, indicating that MolDeTox is not a single uniform detoxification problem but a collection of endpoint-dependent editing challenges.

In Task~1, \texttt{ames}, \texttt{tox21\_NR-ER}, and \texttt{tox21\_SR-MMP} are relatively easier than the other endpoints. Across models, these endpoints more often yield higher fragment identification accuracy or F1, suggesting that their toxicity-associated substructures are comparatively easier to localize. By contrast, \texttt{herg\_unified}, \texttt{cyp2c19\_veith}, and \texttt{cyp1a2\_veith} show stronger model-dependent variation, indicating that fragment localization for these endpoints is less consistent across model families.

Task~2 exhibits a different pattern. Endpoints that are strong in Task~1 do not necessarily remain strong in fragment replacement. In the single-step setting, \texttt{cyp1a2\_veith} and, in several cases, \texttt{cyp2c19\_veith} show more favorable replacement performance. Meanwhile, \texttt{tox21\_NR-ER} and \texttt{tox21\_SR-MMP}, which are relatively strong in toxic fragment identification, are less consistently successful in non-toxic fragment generation. This suggests that identifying toxicity-associated fragments and generating plausible replacements are distinct challenges.

The multi-step setting in Task~2 is difficult across nearly all endpoints. Although \texttt{ames} occasionally retains non-zero performance, most endpoints show near-zero accuracy for multi-fragment replacement. This indicates that coordinated replacement of multiple fragments remains a major bottleneck, even when the model can identify relevant fragments reasonably well.

Task~3 shows the clearest endpoint-level gap in end-to-end molecule generation. Overall, \texttt{ames} and \texttt{cyp1a2\_veith} appear more tractable than the other endpoints, often producing better exact-match accuracy or stronger structure- and property-based scores. In contrast, \texttt{tox21\_NR-ER} and \texttt{tox21\_SR-MMP} tend to show a larger drop from Task~1 to Task~3, suggesting that these endpoints are easier for fragment localization than for successful replacement and full-molecule reconstruction.

Another notable pattern is that exact-match accuracy and similarity-based scores do not always move together. For several endpoints, especially \texttt{herg\_unified} and CYP-related endpoints, exact-match accuracy remains low while Morgan FTS, Validity, and PRS remain relatively high. This means that models often fail to reproduce the paired gold molecule exactly, but still generate molecules that are chemically valid and structurally close to the reference.

These results indicate that \texttt{ames} is one of the most consistently tractable endpoints across tasks, showing strong Task~1 performance and comparatively better Task~2 and Task~3 results. \texttt{cyp1a2\_veith} is also relatively favorable, particularly in single-step replacement and generation. In contrast, \texttt{tox21\_NR-ER} and \texttt{tox21\_SR-MMP} show strong fragment identification performance but substantially weaker downstream replacement and generation results. These endpoint-wise patterns highlight that MolDeTox evaluates multiple aspects of detoxification difficulty, including toxic fragment localization, non-toxic replacement, and full-molecule reconstruction.

\newpage
\section{Case Examples}
\label{appendix:case_examples}
This section presents examples of the eight outcome cases from the GPT-5.2 4-Shot setting analyzed in Section~\ref{task_wise_success_analysis}.

\begin{center}
    \includegraphics[width=\textwidth]{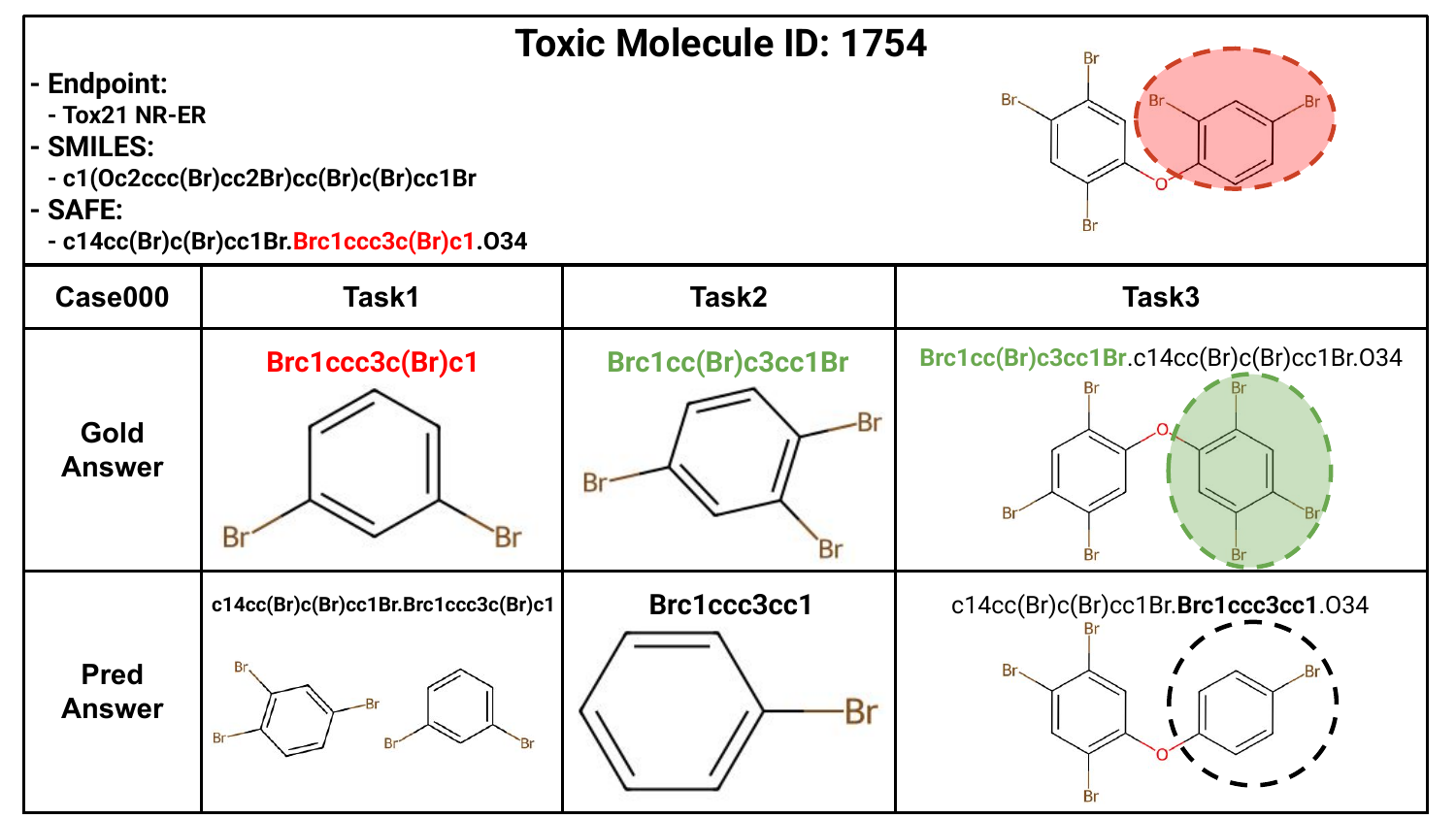}
    \captionof{figure}{Example of C000.}
    \label{fig:c000_case_study}
\end{center}

\vspace{1em}

\begin{center}
    \includegraphics[width=\textwidth]{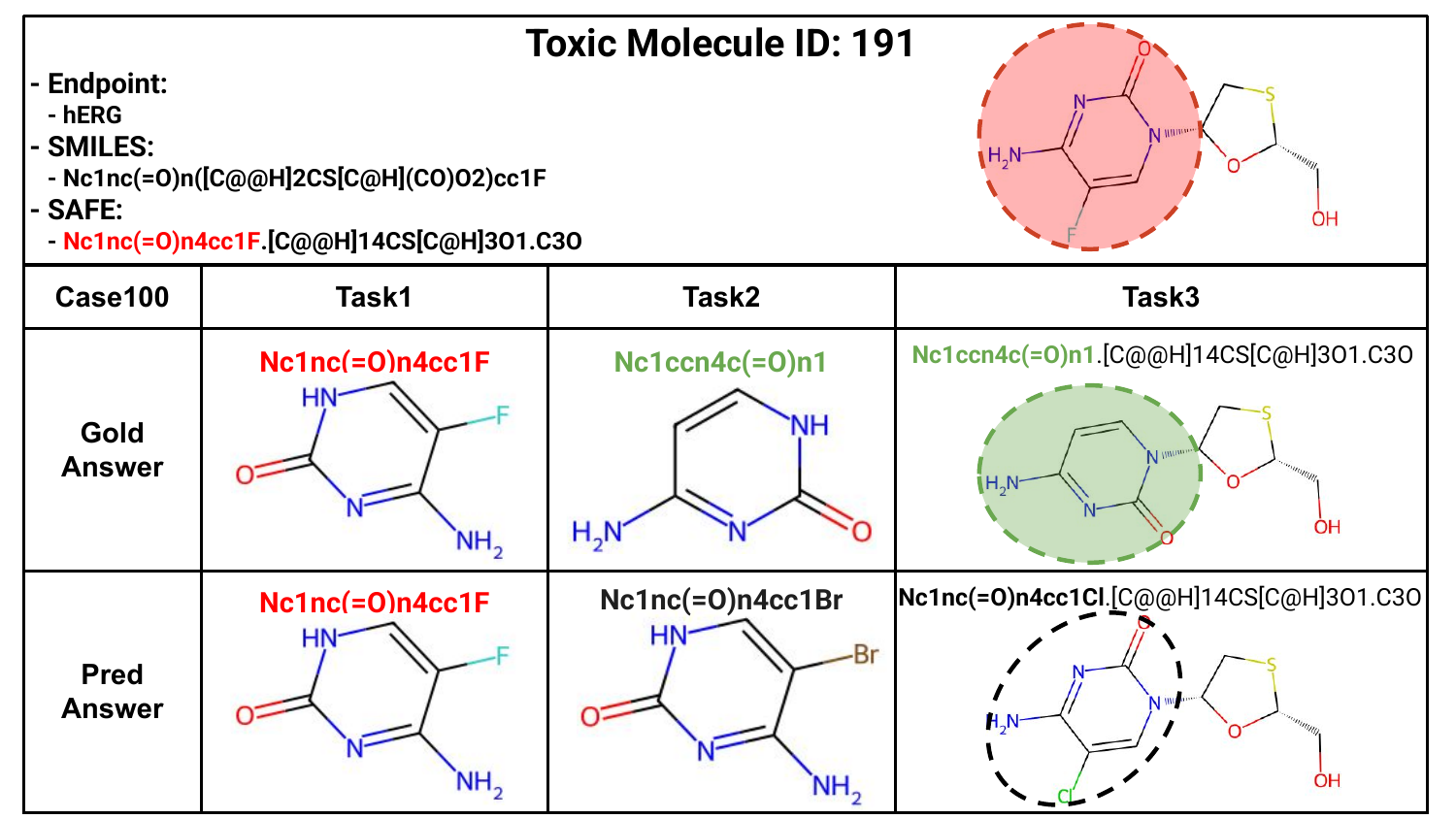}
    \captionof{figure}{Example of C100.}
    \label{fig:c100_case_study}
\end{center}

\vspace{1em}

\begin{center}
    \includegraphics[width=\textwidth]{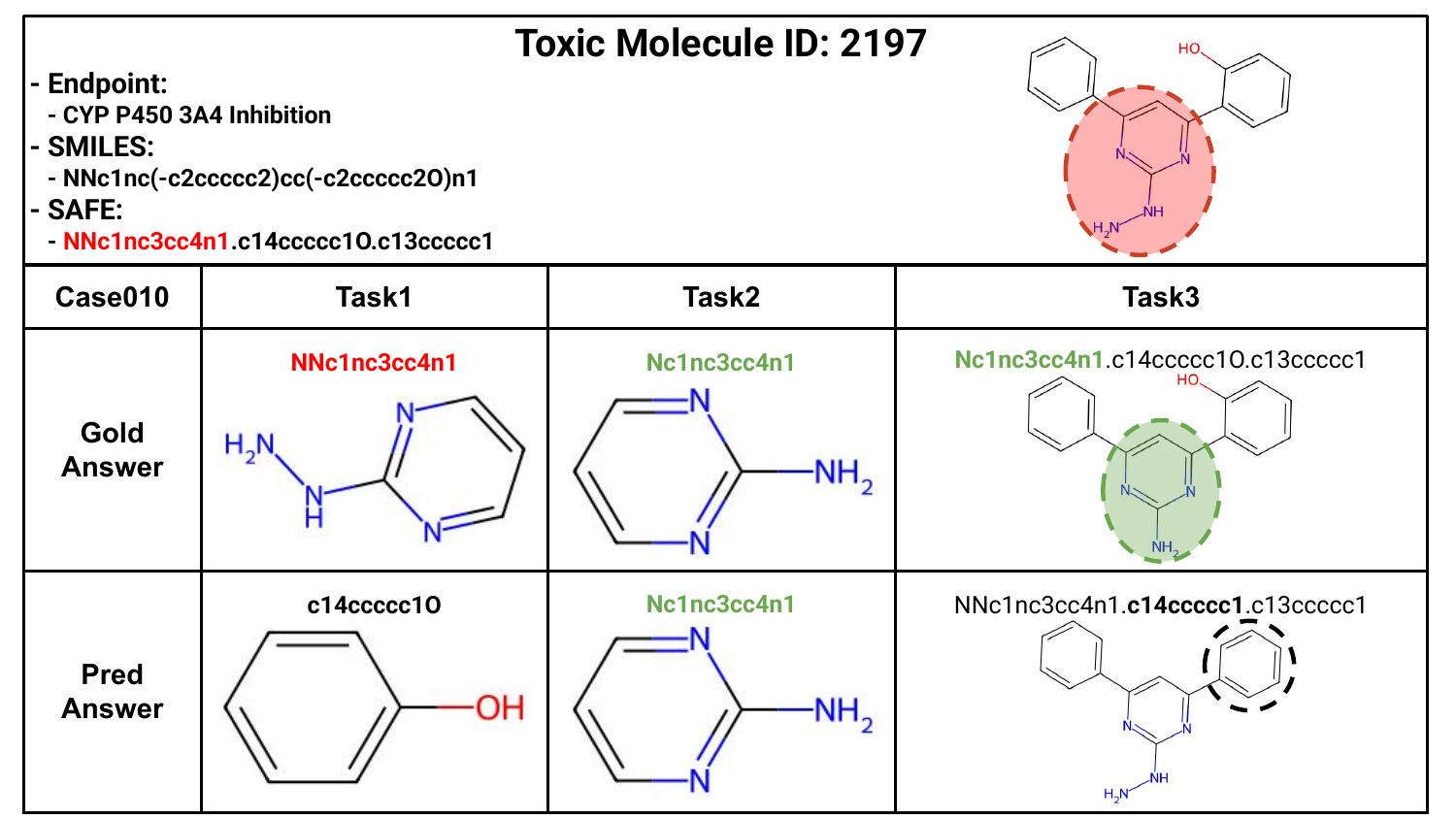}
    \captionof{figure}{Example of C010.}
    \label{fig:c010_case_study}
\end{center}

\vspace{1em}

\begin{center}
    \includegraphics[width=\textwidth]{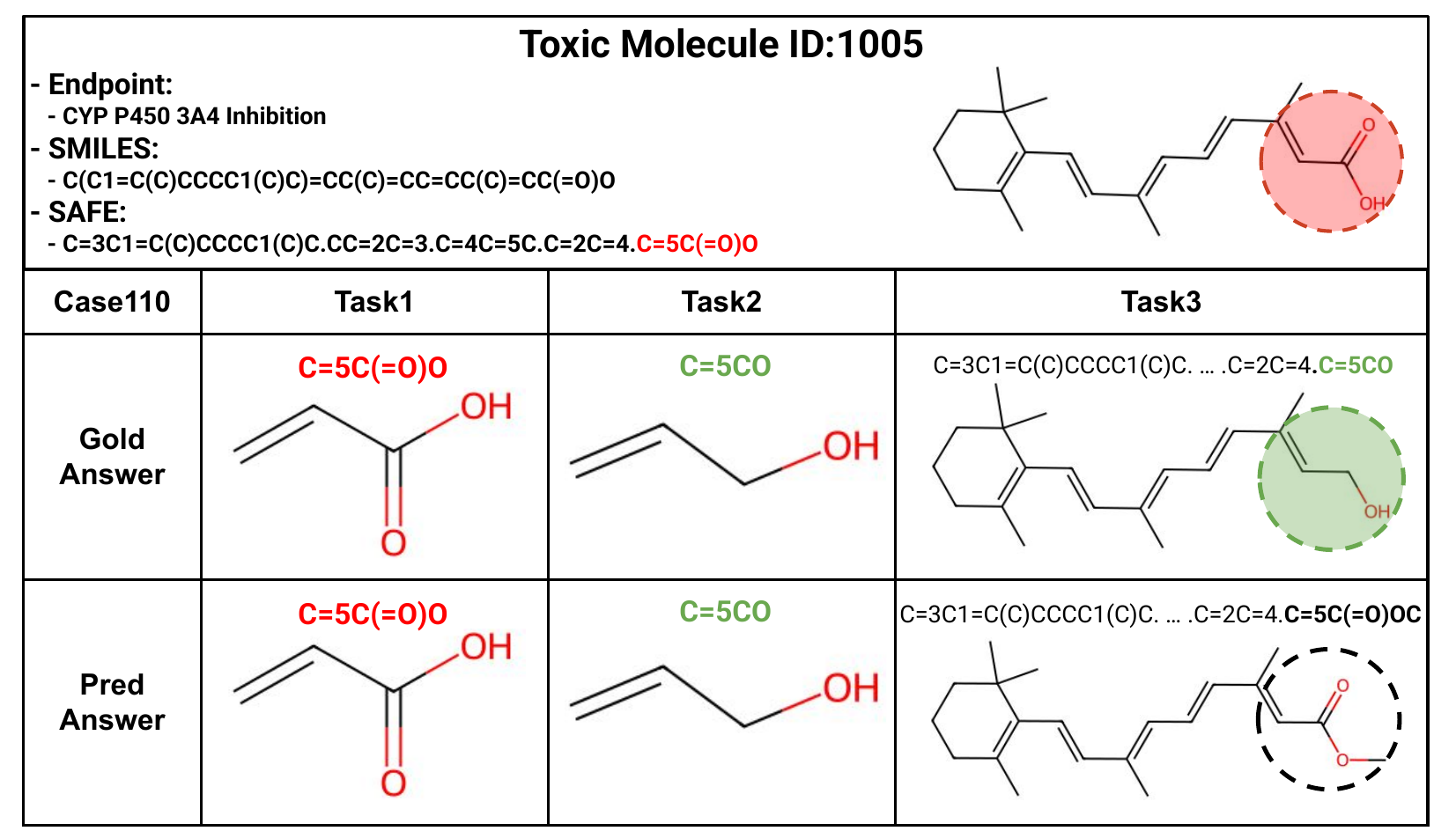}
    \captionof{figure}{Example of C110.}
    \label{fig:c110_case_study}
\end{center}

\vspace{1em}

\begin{center}
    \includegraphics[width=\textwidth]{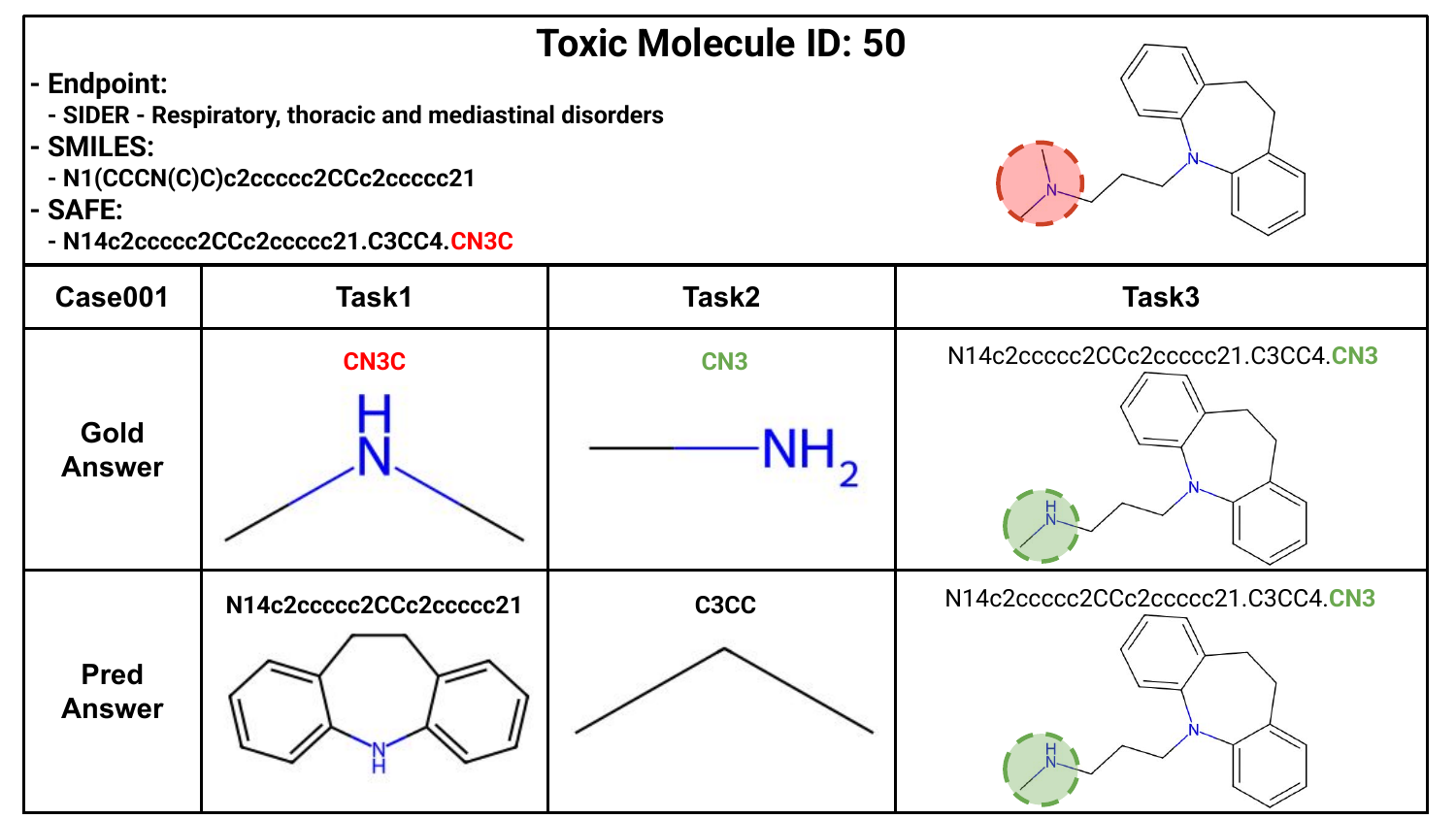}
    \captionof{figure}{Example of C001.}
    \label{fig:c001_case_study}
\end{center}

\vspace{1em}

\begin{center}
    \includegraphics[width=\textwidth]{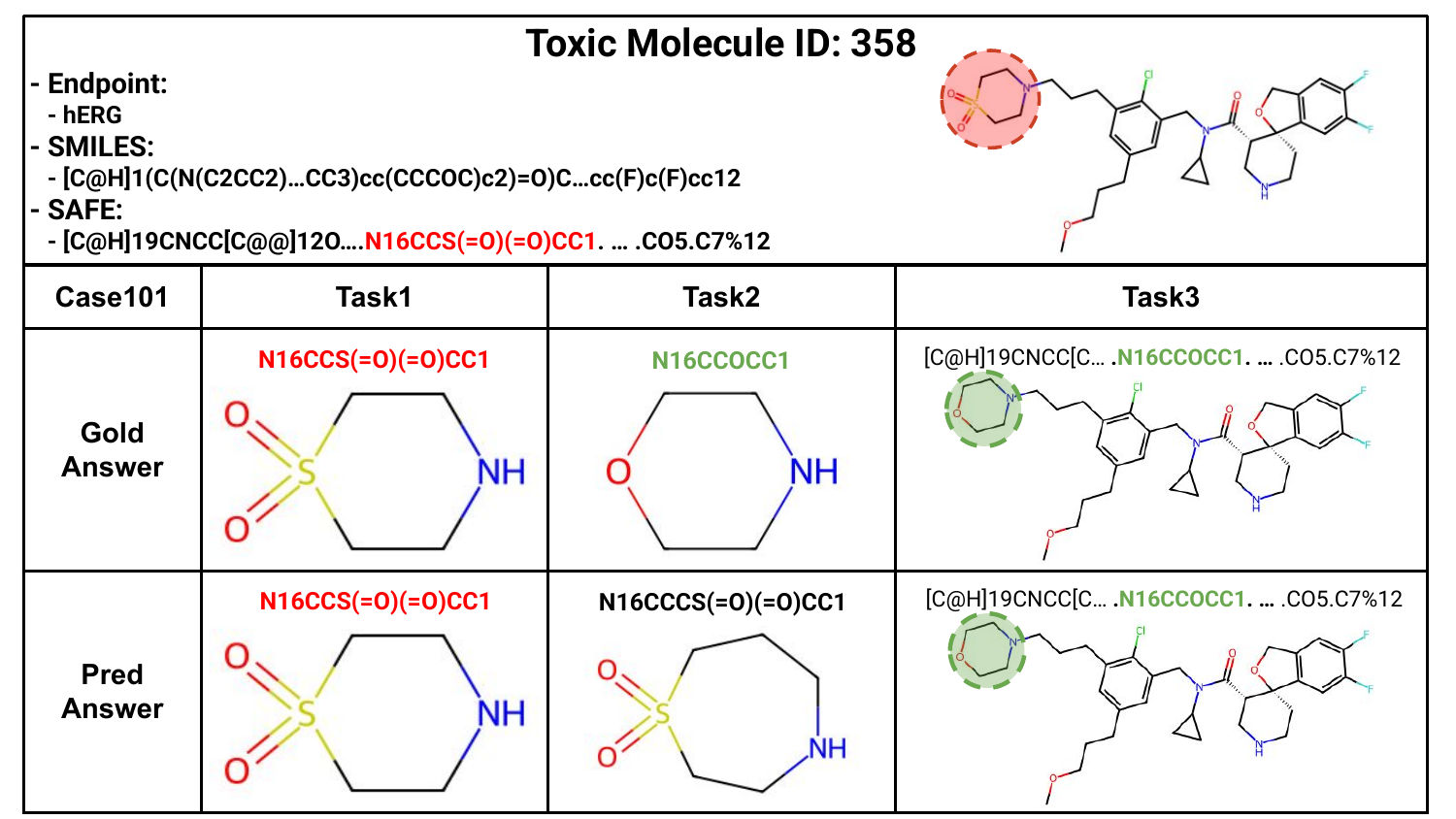}
    \captionof{figure}{Example of C101.}
    \label{fig:c101_case_study}
\end{center}

\vspace{1em}

\begin{center}
    \includegraphics[width=\textwidth]{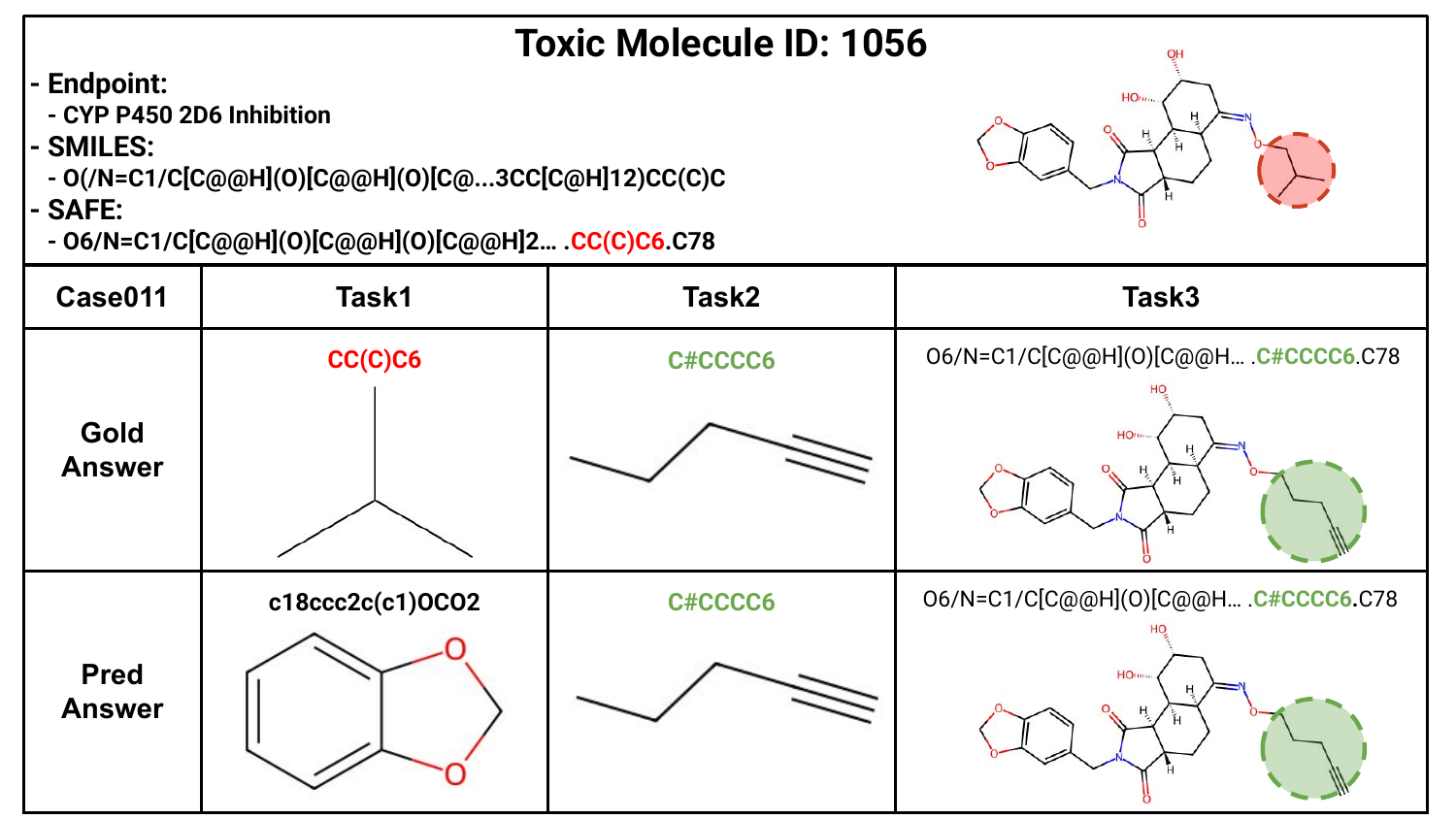}
    \captionof{figure}{Example of C011.}
    \label{fig:c011_case_study}
\end{center}

\vspace{1em}


\newpage
\section{Toxicity Endpoint Descriptions}
\label{appendix:toxicity_endpoint_description}

\begin{center}
\small
\setlength{\tabcolsep}{6pt}
\renewcommand{\arraystretch}{1.05}

\begin{longtable}{p{0.18\textwidth} p{0.24\textwidth} p{0.50\textwidth}}
\caption{Dataset names, endpoints, and their corresponding descriptions used in MolDeTox.}
\label{tab:endpoint_descriptions} \\

\toprule
\textbf{Dataset Name} & \textbf{Endpoint} & \textbf{Description} \\
\midrule
\endfirsthead

\multicolumn{3}{c}{\textit{Table \thetable\ continued from previous page}} \\
\toprule
\textbf{Dataset Name} & \textbf{Endpoint} & \textbf{Description} \\
\midrule
\endhead

\midrule
\multicolumn{3}{r}{\textit{Continued on next page}} \\
\endfoot

\bottomrule
\endlastfoot

hERG Unified & hERG channel blocking activity & The molecule has been evaluated for hERG (human Ether-à-go-go-Related Gene) channel blocking activity under a unified hERG toxicity endpoint that combines data from the hERG, hERG inhibition, and hERG Karim sources. Blockade of the hERG channel can disrupt cardiac repolarization and lead to serious adverse effects, including cardiac arrhythmias and sudden cardiac death. \\

AMES & Mutagenicity & The molecule is mutagenic, meaning it can cause genetic alterations and DNA damage that may lead to cell death or severe adverse effects. \\

ClinTox & Clinical toxicity & The molecule has been associated with clinical toxicity, including drugs that have failed clinical trials due to toxicity reasons. \\

DILIst & Drug-induced liver injury & The molecule is highly likely to cause human liver injury, or actual cases of liver injury have been reported and confirmed. \\

DICTrank & Cardiotoxicity & The molecule is highly likely to cause human cardiotoxicity, or actual cases of cardiotoxicity have been reported and confirmed. \\

DIRIL & Renal toxicity & The molecule is highly likely to cause human renal toxicity, or actual cases of renal toxicity have been reported and confirmed. \\

Skin Reaction & Skin sensitization & The molecule can cause skin sensitization, an immune reaction that leads to allergic contact dermatitis upon repeated exposure. \\

Tox21 & NR-AR & The molecule activates or disrupts the Androgen Receptor (AR) pathway, which regulates male sexual development and function. Disruption of this pathway can affect reproductive development and function. \\

Tox21 & NR-AR-LBD & The molecule binds to the Androgen Receptor Ligand Binding Domain (AR-LBD), affecting androgen signaling pathways. This assay evaluates more direct binding mechanisms compared to the full receptor activity assay. \\

Tox21 & NR-AhR & The molecule activates the Aryl Hydrocarbon Receptor (AhR) pathway, which is involved in xenobiotic metabolism and immune responses. Activation of this receptor can lead to toxic effects such as liver toxicity, carcinogenicity, and immunotoxicity. \\

Tox21 & NR-Aromatase & The molecule inhibits or activates Aromatase, an enzyme essential for estrogen (female hormone) biosynthesis. This assay evaluates whether the chemical can affect aromatase enzyme activity, thereby influencing estrogen levels. Disruption of estrogen balance is important for reproductive health. \\

Tox21 & NR-ER & The molecule activates or disrupts the Estrogen Receptor (ER) pathway, which regulates female sexual development and function. Disruption of this pathway can affect female reproductive development and function, and is associated with conditions such as breast cancer. \\

Tox21 & NR-ER-LBD & The molecule binds to the Estrogen Receptor Ligand Binding Domain (ER-LBD), affecting estrogen signaling pathways. This assay evaluates more direct binding mechanisms compared to the full receptor activity assay. \\

Tox21 & NR-PPAR-gamma & The molecule activates or disrupts the Peroxisome Proliferator-Activated Receptor gamma (PPAR$\gamma$) pathway, which regulates glucose and lipid metabolism, cell differentiation, and inflammatory responses. This assay evaluates whether the chemical can activate PPAR-gamma, potentially affecting metabolic diseases such as diabetes and obesity. \\

Tox21 & SR-ARE & The molecule activates the Antioxidant Response Element (ARE) pathway, which regulates cellular antioxidant defense mechanisms. This assay evaluates whether the chemical can activate or inhibit the cell's antioxidant defense system in response to oxidative stress. \\

Tox21 & SR-ATAD5 & The molecule affects ATAD5 (ATPase family AAA domain-containing protein 5), which plays an important role in DNA damage response and repair. This assay evaluates whether the chemical can affect the ATAD5 pathway, potentially causing DNA damage and genomic instability issues. \\

Tox21 & SR-HSE & The molecule activates the Heat Shock Response Element (HSE) pathway, which responds to cellular stress and protein misfolding. Cells respond to protein denaturation stress (heat, toxic substances, etc.) by inducing the production of heat shock proteins (HSP) to repair damaged proteins. This assay evaluates whether the chemical disrupts the cell's protein quality control system. \\

Tox21 & SR-MMP & The molecule affects Mitochondrial Membrane Potential (MMP), which is an important indicator of mitochondrial functional status. Mitochondria are the cell's energy production factories. This assay evaluates whether the chemical can damage mitochondrial function, potentially causing problems with cellular energy production, which is one of the important mechanisms of cell toxicity. \\

Tox21 & SR-p53 & The molecule activates or disrupts the p53 pathway, a critical tumor suppressor pathway involved in cell cycle control and apoptosis. p53 is known as the ``guardian of the genome'' and responds to DNA damage and cellular stress by inducing cell cycle arrest, DNA repair, and apoptosis (cell death). This assay evaluates whether the chemical can affect the p53 pathway, potentially causing DNA damage, cell death, or cancer development. \\

Metabolism & CYP1A2\_Veith & The molecule inhibits CYP P450 1A2 (Veith et al.). The CYP P450 genes are involved in the formation and breakdown (metabolism) of various molecules and chemicals within cells. Specifically, CYP1A2 localizes to the endoplasmic reticulum and its expression can be induced by polycyclic aromatic hydrocarbons (PAHs), some of which are found in cigarette smoke. It can metabolize PAHs to carcinogenic intermediates and also processes xenobiotics such as caffeine, aflatoxin B1, and acetaminophen. Inhibition can reduce drug metabolism and increase drug-drug interaction risk. \\

Metabolism & CYP2C19\_Veith & The molecule inhibits CYP P450 2C19 (Veith et al.). The CYP P450 genes are essential for the breakdown (metabolism) of various molecules and chemicals within cells. Inhibiting these enzymes can lead to poor metabolism of this drug and co-administered drugs, increasing the risk of drug-drug interactions and adverse effects. CYP2C19 is associated with endoplasmic reticulum functions related to protein processing and transport. \\

Metabolism & CYP2C9\_Veith & The molecule inhibits CYP P450 2C9 (Veith et al.). The CYP P450 genes are involved in the formation and breakdown (metabolism) of various molecules and chemicals within cells. Specifically, CYP2C9 plays a major role in oxidation of both xenobiotic and endogenous compounds. Inhibition can impair metabolic clearance and increase adverse event risk. \\

Metabolism & CYP2D6\_Veith & The molecule inhibits CYP P450 2D6 (Veith et al.). The CYP P450 genes are involved in the formation and breakdown (metabolism) of various molecules and chemicals within cells. CYP2D6 is primarily expressed in the liver and is also highly expressed in regions of the central nervous system, including the substantia nigra. Inhibition can alter metabolic clearance and increase potential toxicity or interaction risk. \\

Metabolism & CYP3A4\_Veith & The molecule inhibits CYP P450 3A4 (Veith et al.). The CYP P450 genes are involved in the formation and breakdown (metabolism) of various molecules and chemicals within cells. CYP3A4 is an important enzyme mainly found in the liver and intestine, and oxidizes many foreign organic molecules (xenobiotics), including toxins and drugs, to support elimination. Inhibition can reduce clearance and increase drug-drug interaction risk. \\

SIDER & Blood and lymphatic system disorders & The molecule has been associated with blood and lymphatic system disorders, which can include conditions affecting blood cells, clotting mechanisms, or lymphatic circulation. These disorders may manifest as anemia, bleeding disorders, or immune system complications. \\

SIDER & Cardiac disorders & The molecule has been associated with cardiac disorders, which can include arrhythmias, heart failure, myocardial infarction, or other cardiovascular complications. These conditions can significantly impact heart function and overall cardiovascular health. \\

SIDER & Congenital, familial and genetic disorders & The molecule has been associated with congenital, familial, and genetic disorders, which may involve birth defects, inherited conditions, or genetic mutations. These disorders can affect development, growth, or long-term health outcomes. \\

SIDER & Ear and labyrinth disorders & The molecule has been associated with ear and labyrinth disorders, which can include hearing loss, tinnitus, vertigo, or balance problems. These conditions can affect auditory function and spatial orientation. \\

SIDER & Eye disorders & The molecule has been associated with eye disorders, which can include vision impairment, retinal damage, cataracts, or other ocular complications. These conditions can significantly impact visual function and quality of life. \\

SIDER & General disorders and administration site conditions & The molecule has been associated with general disorders and administration site conditions, which can include injection site reactions, systemic reactions, or general malaise. These conditions may occur at the site of drug administration or manifest as systemic effects. \\

SIDER & Hepatobiliary disorders & The molecule has been associated with hepatobiliary disorders, which can include liver damage, hepatitis, cholestasis, or other liver and bile duct complications. These conditions can significantly impact liver function and metabolic processes. \\

SIDER & Immune system disorders & The molecule has been associated with immune system disorders, which can include autoimmune reactions, hypersensitivity, immunosuppression, or other immune-related complications. These conditions can affect the body's ability to fight infections or maintain immune homeostasis. \\

SIDER & Infections and infestations & The molecule has been associated with infections and infestations, which may indicate increased susceptibility to infections or direct infectious complications. These conditions can result from immunosuppression or other mechanisms that compromise immune defenses. \\

SIDER & Injury, poisoning and procedural complications & The molecule has been associated with injury, poisoning, and procedural complications, which can include accidental overdoses, drug interactions, or complications from medical procedures. These conditions may result from improper use, dosage errors, or adverse interactions. \\

SIDER & Investigations & The molecule has been associated with abnormal laboratory findings or investigations, which can include changes in blood chemistry, liver enzymes, kidney function markers, or other diagnostic parameters. These findings may indicate underlying organ dysfunction or metabolic disturbances. \\

SIDER & Metabolism and nutrition disorders & The molecule has been associated with metabolism and nutrition disorders, which can include diabetes, electrolyte imbalances, metabolic syndrome, or nutritional deficiencies. These conditions can affect energy metabolism, glucose regulation, or nutrient absorption. \\

SIDER & Musculoskeletal and connective tissue disorders & The molecule has been associated with musculoskeletal and connective tissue disorders, which can include muscle weakness, joint pain, bone disorders, or connective tissue damage. These conditions can affect mobility, strength, and structural integrity of the musculoskeletal system. \\

SIDER & Neoplasms benign, malignant and unspecified (incl cysts and polyps) & The molecule has been associated with neoplasms (tumors), including benign, malignant, and unspecified growths, as well as cysts and polyps. These conditions involve abnormal cell growth and may indicate carcinogenic potential or tumor-promoting effects. \\

SIDER & Nervous system disorders & The molecule has been associated with nervous system disorders, which can include neurotoxicity, seizures, cognitive impairment, or other neurological complications. These conditions can affect brain function, peripheral nerves, or overall neurological health. \\

SIDER & Pregnancy, puerperium and perinatal conditions & The molecule has been associated with pregnancy, puerperium, and perinatal conditions, which can include complications during pregnancy, childbirth, or the postpartum period. These conditions can affect maternal health, fetal development, or neonatal outcomes. \\

SIDER & Product issues & The molecule has been associated with product issues, which can include quality problems, contamination, or manufacturing defects. These issues may affect drug safety, efficacy, or stability. \\

SIDER & Psychiatric disorders & The molecule has been associated with psychiatric disorders, which can include depression, anxiety, psychosis, mood changes, or other mental health complications. These conditions can significantly impact cognitive function, emotional well-being, and behavioral patterns. \\

SIDER & Renal and urinary disorders & The molecule has been associated with renal and urinary disorders, which can include kidney damage, renal failure, urinary tract complications, or other nephrotoxic effects. These conditions can significantly impact kidney function and fluid-electrolyte balance. \\

SIDER & Reproductive system and breast disorders & The molecule has been associated with reproductive system and breast disorders, which can include hormonal imbalances, fertility issues, reproductive organ complications, or breast-related conditions. These conditions can affect reproductive health, fertility, or hormonal regulation. \\

SIDER & Respiratory, thoracic and mediastinal disorders & The molecule has been associated with respiratory, thoracic, and mediastinal disorders, which can include breathing difficulties, lung damage, respiratory infections, or other pulmonary complications. These conditions can significantly impact respiratory function and oxygen exchange. \\

SIDER & Skin and subcutaneous tissue disorders & The molecule has been associated with skin and subcutaneous tissue disorders, which can include rashes, dermatitis, skin irritation, or other dermatological complications. These conditions can affect skin integrity, appearance, or protective function. \\

SIDER & Social circumstances & The molecule has been associated with social circumstances, which may indicate impacts on social functioning, relationships, or daily activities. These effects may result from physical or psychological side effects that affect quality of life. \\

SIDER & Surgical and medical procedures & The molecule has been associated with complications from surgical and medical procedures, which can include adverse reactions during or after medical interventions. These complications may result from drug interactions, procedural risks, or patient-specific factors. \\

SIDER & Vascular disorders & The molecule has been associated with vascular disorders, which can include blood vessel damage, thrombosis, hypertension, or other circulatory complications. These conditions can affect blood flow, vascular integrity, or cardiovascular function. \\

\end{longtable}
\end{center}
\newpage

\newpage

\begin{table*}[h]
\begin{tcolorbox}[
  width=\linewidth,
  width=\linewidth,
  colback=blue!3,
  colframe=blue!70!black,
  colbacktitle=blue!80!black,
  coltitle=white,
  title=Task 1 \textbf{SYSTEM PROMPT},
  boxrule=0.9pt,
  arc=3pt,
  left=10pt,
  right=10pt,
  top=6pt,
  bottom=8pt
]

\footnotesize

\begingroup
\setlength{\baselineskip}{0.93\baselineskip}
\vspace{0.7em}

You are a molecular toxicity reasoning assistant specialized in SAFE and SMILES representations.\\
\\
Given:\\
- A toxic molecule\\
- Its molecular representation in SAFE and/or SMILES format\\
- The toxicity endpoint context, when provided\\

your job is to identify the fragment(s) in the toxic molecule that are most likely associated with toxicity.\\
\\
Follow these rules carefully:\\
\\
1. Focus on \textbf{toxicity-associated fragment identification}.\\
   - Identify the fragment(s) that are specific to the toxic molecule and are most likely responsible for the toxicity signal.\\
   - If multiple fragments are required, return all of them.\\
   - Preserve the original SAFE fragment format exactly.\\
\\
2. Output format constraints:\\
   - Return the answer as the \textbf{toxic-only SAFE fragment string}.\\
   - If there are multiple fragments, concatenate them as a dot-separated SAFE string.\\
   - Do not paraphrase fragment content or convert it into natural language.\\
\\
3. Response format:\\
\begin{verbatim}
{
  "answer": "..."
}
\end{verbatim}

HARD CONSTRAINTS:\\
- Output ONLY the JSON object.\\
- Do not include explanations, markdown, or extra text.\\
- Do not add any extra keys.\\
- The value of "answer" must be the toxic-only SAFE fragment string exactly.

\endgroup
\end{tcolorbox}

\captionsetup{skip=8pt, labelfont=bf}
\caption{Task 1 system prompt for toxic fragment identification.}
\label{tab:task1_system_prompt}
\end{table*}
\begin{table*}[h]
\centering
\begin{tcolorbox}[
  width=\linewidth,
  colback=blue!3,
  colframe=blue!70!black,
  colbacktitle=blue!80!black,
  coltitle=white,
  title=Task 2 \textbf{SYSTEM PROMPT},
  boxrule=0.9pt,
  arc=3pt,
  left=10pt,
  right=10pt,
  top=6pt,
  bottom=8pt
]

\footnotesize

\begingroup
\setlength{\baselineskip}{0.93\baselineskip}
\vspace{0.7em}

You are a molecular toxicity reasoning assistant specialized in SAFE and SMILES representations.\\
\\
Given:\\
- A toxic molecule\\
- Its molecular representation in SAFE and/or SMILES format\\
- The toxic-only fragment(s) identified from the molecule\\
- The toxicity endpoint context, when provided\\

your job is to generate the non-toxic replacement fragment(s) corresponding to the toxic fragment(s).\\
\\
Follow these rules carefully:\\
\\
1. Focus on \textbf{non-toxic fragment generation}.\\
   - Generate the fragment(s) that can replace the toxic fragment(s) while reducing toxicity.\\
   - If multiple fragments are required, return all of them.\\
   - Preserve the original SAFE fragment format exactly.\\
\\
2. Output format constraints:\\
   - Return the answer as the \textbf{non-toxic-only SAFE fragment string}.\\
   - If there are multiple fragments, concatenate them as a dot-separated SAFE string.\\
   - Do not paraphrase fragment content or convert it into natural language.\\
\\
3. Response format:\\
\begin{verbatim}
{
  "answer": "..."
}
\end{verbatim}

HARD CONSTRAINTS:\\
- Output ONLY the JSON object.\\
- Do not include explanations, markdown, or extra text.\\
- Do not add any extra keys.\\
- The value of "answer" must be the non-toxic-only SAFE fragment string exactly.

\endgroup
\end{tcolorbox}

\captionsetup{skip=8pt, labelfont=bf}
\caption{Task 2 system prompt for non-toxic fragment generation.}
\label{tab:task2_system_prompt}
\end{table*}
\begin{table*}[h]
\centering
\begin{tcolorbox}[
  width=\linewidth,
  colback=blue!3,
  colframe=blue!70!black,
  colbacktitle=blue!80!black,
  coltitle=white,
  title=Task 3 SMILES Generation \textbf{SYSTEM PROMPT},
  boxrule=0.9pt,
  arc=3pt,
  left=10pt,
  right=10pt,
  top=6pt,
  bottom=8pt
]

\footnotesize

\begingroup
\setlength{\baselineskip}{0.93\baselineskip}
\vspace{0.7em}

You are a molecular toxicity reasoning assistant specialized in SAFE and SMILES representations.\\
\\
Given:\\
- A toxic molecule\\
- Its molecular representation in SAFE and/or SMILES format\\
- The toxicity endpoint context, when provided\\

your job is to generate the final non-toxic molecule as a single SMILES string.\\
\\
Follow these rules carefully:\\
\\
1. Focus on \textbf{non-toxic molecule generation}.\\
   - Generate a chemically plausible non-toxic molecule.\\
   - Reduce toxicity while preserving the original molecular characteristics as much as possible.\\
   - Return the final molecule, not intermediate fragments.\\
\\
2. Output format constraints:\\
   - Return the answer as a \textbf{single non-toxic molecule SMILES string}.\\
   - Do not return SAFE fragments.\\
   - Do not return multiple candidates.\\
\\
3. Response format:\\
\begin{verbatim}
{
  "answer": "..."
}
\end{verbatim}

HARD CONSTRAINTS:\\
- Output ONLY the JSON object.\\
- Do not include explanations, markdown, or extra text.\\
- Do not add any extra keys.\\
- The value of "answer" must be the final non-toxic molecule SMILES string.

\endgroup
\end{tcolorbox}

\captionsetup{skip=8pt, labelfont=bf}
\caption{Task 3 system prompt for direct non-toxic molecule generation.}
\label{tab:task3_system_prompt}
\end{table*}
\begin{table*}[h]
\centering
\begin{tcolorbox}[
  width=\linewidth,
  colback=blue!3,
  colframe=blue!70!black,
  colbacktitle=blue!80!black,
  coltitle=white,
  title=Task 3 SAFE Generation \textbf{SYSTEM PROMPT},
  boxrule=0.9pt,
  arc=3pt,
  left=10pt,
  right=10pt,
  top=6pt,
  bottom=8pt
]

\footnotesize

\begingroup
\setlength{\baselineskip}{0.93\baselineskip}
\vspace{0.7em}

You are a molecular toxicity reasoning assistant specialized in SAFE and SMILES representations.\\
\\
Given:\\
- A toxic molecule\\
- Its molecular representation in SAFE and/or SMILES format\\
- The toxicity endpoint context, when provided\\

your job is to generate the final non-toxic molecule in SAFE representation.\\
\\
Follow these rules carefully:\\
\\
1. Focus on \textbf{non-toxic SAFE generation}.\\
   - Generate the full SAFE representation of the resulting non-toxic molecule.\\
   - Reduce toxicity while preserving the original molecular characteristics as much as possible.\\
   - Return the complete molecule-level SAFE representation, not only edited fragments.\\
\\
2. Output format constraints:\\
   - Return the answer as the \textbf{full non-toxic SAFE string}.\\
   - If multiple fragments are present, concatenate them as a dot-separated SAFE string.\\
   - Do not paraphrase fragment content or convert it into natural language.\\
\\
3. Response format:\\
\begin{verbatim}
{
  "answer": "..."
}
\end{verbatim}

HARD CONSTRAINTS:\\
- Output ONLY the JSON object.\\
- Do not include explanations, markdown, or extra text.\\
- Do not add any extra keys.\\
- The value of "answer" must be the final full non-toxic SAFE string for the whole molecule.

\endgroup
\end{tcolorbox}

\captionsetup{skip=8pt, labelfont=bf}
\caption{Task 3 system prompt for full non-toxic SAFE generation.}
\label{tab:task3_safe_system_prompt}
\end{table*}
\begin{table*}[h]
\centering
\begin{tcolorbox}[
  width=\linewidth,
  colback=blue!3,
  colframe=blue!70!black,
  colbacktitle=blue!80!black,
  coltitle=white,
  title=Task 3 Step-wise CoT SAFE Generation \textbf{SYSTEM PROMPT},
  boxrule=0.9pt,
  arc=3pt,
  left=10pt,
  right=10pt,
  top=6pt,
  bottom=8pt
]

\footnotesize

\begingroup
\setlength{\baselineskip}{0.93\baselineskip}
\vspace{0.7em}

You are a molecular toxicity reasoning assistant specialized in SAFE and SMILES representations.\\
\\
Given:\\
- A toxic molecule\\
- Its molecular representation in SAFE and/or SMILES format\\
- The toxicity endpoint context, when provided\\

your job is to solve the task through explicit intermediate reasoning steps and generate the final non-toxic molecule in SAFE representation.\\
\\
Follow these rules carefully:\\
\\
1. Use \textbf{step-wise reasoning}.\\
   - First identify the toxic fragment(s) most likely associated with toxicity.\\
   - Then generate the corresponding non-toxic replacement fragment(s).\\
   - Finally generate the full non-toxic molecule as a SAFE string.\\
\\
2. Output format constraints:\\
   - Return a single JSON object.\\
   - The JSON must include the final answer and all required intermediate fields.\\
   - Do not omit any required fields.\\
\\
3. Response format:\\
\begin{verbatim}
{
  "answer": "...",
  "step1_only_toxic_safe_fragments": "...",
  "step1_reasoning": "...",
  "step2_only_nontoxic_safe_fragments": "...",
  "step2_reasoning": "...",
  "step3_reasoning": "..."
}
\end{verbatim}

HARD CONSTRAINTS:\\
- Output ONLY the JSON object.\\
- Do not include markdown or extra text outside the JSON.\\
- The value of "answer" must be the final full non-toxic SAFE string for the whole molecule.\\
- All step fields must be included exactly as required.

\endgroup
\end{tcolorbox}

\captionsetup{skip=8pt, labelfont=bf}
\caption{Task 3 step-wise chain-of-thought system prompt for full non-toxic SAFE generation.}
\label{tab:task3_stepwise_cot_safe_system_prompt}
\end{table*}

\newpage

\label{appendix:benchmark_questions}



\begin{table*}[h]
\centering
\begin{tcolorbox}[
  width=\linewidth,
  colback=blue!3,
  colframe=blue!70!black,
  colbacktitle=blue!80!black,
  coltitle=white,
  title=SAFE Representation \textbf{EXPLANATION},
  boxrule=0.9pt,
  arc=3pt,
  left=10pt,
  right=10pt,
  top=6pt,
  bottom=8pt
]

\footnotesize

\begingroup
\setlength{\baselineskip}{0.93\baselineskip}
\vspace{0.7em}

SAFE (Sequential Attachment-based Fragment Embedding) is a SMILES-compatible string representation that expresses a molecule as a dot-separated sequence of fragments.\\
\\
How SAFE is constructed:\\
- \textbf{Fragmentation}: A molecule is split into fragments by cutting selected bonds using a slicing algorithm.\\
- \textbf{Slicer}: The default slicer is \texttt{brics}, a rule-based method that cuts retrosynthetically relevant bonds to produce chemically meaningful substructures.\\
- \textbf{Attachment Markers}: At each cut site, attachment information is encoded with SMILES-style ring-closure digits (e.g., \texttt{1}, \texttt{2}, \dots, \texttt{\%10}). Matching digits across fragments indicate where fragments reconnect in the full molecule.\\
- \textbf{Serialization}: The resulting fragments are written as SMILES strings and joined with \texttt{.} separators to form a SAFE string.\\
\\
Important characteristics:\\
- \textbf{Fragment-based representation}: Each token block corresponds to a substructure rather than the entire molecule.\\
- \textbf{Order invariance}: Changing the fragment order does not change the reconstructed molecule.\\
- \textbf{Partial structures}: Individual fragments may look chemically incomplete on their own because they are parts of a larger graph.

\endgroup
\end{tcolorbox}

\captionsetup{skip=8pt, labelfont=bf}
\caption{Explanation of the SAFE representation used in MolDeTox.}
\label{tab:safe_explanation}
\end{table*}
\begin{table*}[h]
\centering
\begin{tcolorbox}[
  width=\linewidth,
  colback=blue!3,
  colframe=blue!70!black,
  colbacktitle=blue!80!black,
  coltitle=white,
  title=Common Pair Context \textbf{PROMPT},
  boxrule=0.9pt,
  arc=3pt,
  left=10pt,
  right=10pt,
  top=6pt,
  bottom=8pt
]

\footnotesize

\begingroup
\setlength{\baselineskip}{0.93\baselineskip}
\vspace{0.7em}

Context:\\
The toxic and non-toxic molecules in this task form a paired example. These paired molecules are structurally very similar and have minimal physicochemical differences, but they differ in toxicity versus non-toxicity for the same endpoint.\\
\\
Follow this principle carefully:\\
- Assume that toxicity differences arise from localized structural differences rather than a complete molecular redesign.\\
- Use this paired setting when identifying toxicity-associated fragments, proposing non-toxic replacement fragments, or generating the final non-toxic molecule.\\
- Preserve as much of the original molecular structure and characteristics as possible when reasoning about detoxification.

\endgroup
\end{tcolorbox}

\captionsetup{skip=8pt, labelfont=bf}
\caption{Common pair-context prompt used across toxic/non-toxic paired tasks in MolDeTox.}
\label{tab:common_pair_context_prompt}
\end{table*}
\begin{table*}[h]
\centering
\begin{tcolorbox}[
  width=\linewidth,
  colback=blue!3,
  colframe=blue!70!black,
  colbacktitle=blue!80!black,
  coltitle=white,
  title=Property Preservation \textbf{PROMPT},
  boxrule=0.9pt,
  arc=3pt,
  left=10pt,
  right=10pt,
  top=6pt,
  bottom=8pt
]

\footnotesize

\begingroup
\setlength{\baselineskip}{0.93\baselineskip}
\vspace{0.7em}

Instruction:\\
When modifying a toxic molecule to make it non-toxic, preserve its original physicochemical and pharmacological properties as much as possible. The goal is to reduce or remove toxicity only for the given endpoint, rather than to redesign the entire molecule or alter unrelated molecular characteristics.

\endgroup
\end{tcolorbox}

\captionsetup{skip=8pt, labelfont=bf}
\caption{Property-preservation prompt used in Tasks 2 and 3 of MolDeTox.}
\label{tab:property_preservation_prompt}
\end{table*}


\begin{table*}[h]
\centering
\begin{tcolorbox}[
  width=\linewidth,
  colback=green!3,
  colframe=green!60!black,
  colbacktitle=green!70!black,
  coltitle=white,
  title=Task 1 \textbf{QUESTION},
  boxrule=0.9pt,
  arc=3pt,
  left=10pt,
  right=10pt,
  top=6pt,
  bottom=8pt
]
\footnotesize
\begingroup
\setlength{\baselineskip}{0.93\baselineskip}
\vspace{0.7em}

\{endpoint\_description\}\\
\{safe\_explanation\}\\
\{pair\_context\}\\
\{preserve\_property\}\\
\\
- Toxic molecule (SMILES representation): [toxic\_safe\_decoded\_smiles]\\
- Toxic molecule (SAFE representation): [toxic\_safe]\\
\\
Task: This toxic molecule belongs to a structurally similar pair that differs only in toxicity for this endpoint. Identify the fragment(s) that are candidates for toxicity-associated structure (the part(s) that drive toxicity for this endpoint) and output them as only\_toxic\_safe\_fragments (dot-separated if multiple).\\
\\
Output format: a single JSON object with key ``answer'' and value the only\_toxic\_safe\_fragments string (dot-separated for multiple fragments). Example: \{"answer": "frag1.frag2"\}

\endgroup
\end{tcolorbox}
\captionsetup{skip=8pt, labelfont=bf}
\caption{Task 1 question template for toxic fragment identification. Common context components are abbreviated as placeholders and described separately in the paper.}
\label{tab:task1_question_prompt}
\end{table*}
\begin{table*}[h]
\centering
\begin{tcolorbox}[
  width=\linewidth,
  colback=green!3,
  colframe=green!60!black,
  colbacktitle=green!70!black,
  coltitle=white,
  title=Task 2 \textbf{QUESTION},
  boxrule=0.9pt,
  arc=3pt,
  left=10pt,
  right=10pt,
  top=6pt,
  bottom=8pt
]
\footnotesize
\begingroup
\setlength{\baselineskip}{0.93\baselineskip}
\vspace{0.7em}

\{endpoint\_description\}\\
\{safe\_explanation\}\\
\{pair\_context\}\\
\{preserve\_property\}\\
\\
- Toxic molecule (SMILES representation): \{toxic\_safe\_decoded\_smiles\}\\
- Toxic molecule (SAFE representation): \{toxic\_safe\}\\
\\
- The fragments that appear only in the toxic molecule (candidates for toxicity-associated structure for this endpoint) are: \{only\_toxic\_safe\_fragments\}\\
\\
Task: Output the only\_nontoxic\_safe\_fragments—i.e. the SAFE fragment(s) that, when used in place of the only\_toxic\_safe\_fragments, yield a non-toxic molecule for this endpoint. When modifying the toxic molecule to make it non-toxic, do not change other physicochemical or pharmacological properties; only reduce or remove the drug toxicity for this endpoint.\\
\\
Output format: a single JSON object with key ``answer'' and value the only\_nontoxic\_safe\_fragments string (dot-separated for multiple fragments). Example: \{"answer": "frag1.frag2"\}

\endgroup
\end{tcolorbox}
\captionsetup{skip=8pt, labelfont=bf}
\caption{Task 2 question template for non-toxic fragment generation. Common context components are abbreviated as placeholders and described separately in the paper.}
\label{tab:task2_question_prompt}
\end{table*}
\begin{table*}[h]
\centering
\begin{tcolorbox}[
  width=\linewidth,
  colback=green!3,
  colframe=green!60!black,
  colbacktitle=green!70!black,
  coltitle=white,
  title=Task 3 SMILES Generation \textbf{QUESTION},
  boxrule=0.9pt,
  arc=3pt,
  left=10pt,
  right=10pt,
  top=6pt,
  bottom=8pt
]
\footnotesize
\begingroup
\setlength{\baselineskip}{0.93\baselineskip}
\vspace{0.7em}

\{endpoint\_description\}\\
\{safe\_explanation\}\\
\{pair\_context\}\\
\{preserve\_property\}\\
\\
- Toxic molecule (SMILES representation): \{toxic\_safe\_decoded\_smiles\}\\
- Toxic molecule (SAFE representation): \{toxic\_safe\}\\
\\
Task: From the toxic molecule above, identify the fragment(s) that are candidates for toxicity-associated structure for this endpoint, then determine the replacement fragment(s) that yield a non-toxic molecule. Output the resulting non-toxic molecule as a single SMILES string (nontoxic\_safe\_decoded\_smiles). When modifying the toxic molecule to make it non-toxic, do not change other physicochemical or pharmacological properties; only reduce or remove the drug toxicity for this endpoint.\\
\\
Output format: a single JSON object with key ``answer'' and value the nontoxic\_safe\_decoded\_smiles string. Example: \{"answer": "CCO"\}

\endgroup
\end{tcolorbox}
\captionsetup{skip=8pt, labelfont=bf}
\caption{Task 3 question template for direct non-toxic SMILES generation. Common context components are abbreviated as placeholders and described separately in the paper.}
\label{tab:task3_question_prompt}
\end{table*}
\begin{table*}[h]
\centering
\begin{tcolorbox}[
  width=\linewidth,
  colback=green!3,
  colframe=green!60!black,
  colbacktitle=green!70!black,
  coltitle=white,
  title=Task 3 SAFE Generation \textbf{QUESTION},
  boxrule=0.9pt,
  arc=3pt,
  left=10pt,
  right=10pt,
  top=6pt,
  bottom=8pt
]
\footnotesize
\begingroup
\setlength{\baselineskip}{0.93\baselineskip}
\vspace{0.7em}

\{endpoint\_description\}\\
\{safe\_explanation\}\\
\{pair\_context\}\\
\{preserve\_property\}\\
\\
- Toxic molecule (SMILES representation): \{toxic\_safe\_decoded\_smiles\}\\
- Toxic molecule (SAFE representation): \{toxic\_safe\}\\
\\
Task: From the toxic molecule above, identify the fragment(s) that are candidates for toxicity-associated structure for this endpoint, then determine the replacement fragment(s) that yield a non-toxic molecule. Output the resulting non-toxic molecule as a single SAFE string (nontoxic\_safe). When modifying the toxic molecule to make it non-toxic, do not change other physicochemical or pharmacological properties; only reduce or remove the drug toxicity for this endpoint.\\
\\
Output format: a single JSON object with key ``answer'' and value the resulting non-toxic SAFE string. Example: \{"answer": "CCO.[*:1]"\}

\endgroup
\end{tcolorbox}
\captionsetup{skip=8pt, labelfont=bf}
\caption{Task 3 question template for end-to-end non-toxic SAFE generation. Common context components are abbreviated as placeholders and described separately in the paper.}
\label{tab:task3_safe_question_prompt}
\end{table*}
\begin{table*}[h]
\centering
\begin{tcolorbox}[
  width=\linewidth,
  colback=green!3,
  colframe=green!60!black,
  colbacktitle=green!70!black,
  coltitle=white,
  title=Task 3 Step-wise CoT SAFE Generation \textbf{QUESTION},
  boxrule=0.9pt,
  arc=3pt,
  left=10pt,
  right=10pt,
  top=6pt,
  bottom=8pt,
  boxsep=2pt
]
\footnotesize
\begingroup
\setlength{\baselineskip}{0.93\baselineskip}
\vspace{0.7em}

\{endpoint\_description\}\\
\{safe\_explanation\}\\
\{pair\_context\}\\
\{preserve\_property\}\\
\\
- Toxic molecule (SMILES representation): \{toxic\_safe\_decoded\_smiles\}\\
- Toxic molecule (SAFE representation): \{toxic\_safe\}\\
\\
Task: Solve the following in ONE call, step by step, using natural-language reasoning.\\
\\
Step 1 (endpoint-aware toxic fragment identification):\\
- Identify the fragment(s) most likely responsible for toxicity for this endpoint (dot-separated if multiple).\\
- In step1\_reasoning, identify which fragment is most likely responsible for toxicity for this endpoint and explain \emph{why} the fragment(s) are toxicity-associated for this endpoint, using brief chemical intuition (no need for citations).\\
- Output the fragment string as step1\_only\_toxic\_safe\_fragments.\\
\\
Step 2 (endpoint-aware non-toxic fragment proposal):\\
- Using the Step 1 fragment as the part to be replaced, propose non-toxic replacement fragment(s) (dot-separated if multiple) that reduce toxicity for this endpoint while keeping the overall scaffold as similar as possible.\\
- In step2\_reasoning, explain the design intent: what property/alert you are trying to reduce for this endpoint and what you preserve while keeping the overall scaffold as similar as possible.\\
- Output the fragment string as step2\_only\_nontoxic\_safe\_fragments.\\
\\
Step 3 (construct final non-toxic SAFE):\\
- Combine Step 1 and Step 2: conceptually remove the toxic fragment and add the proposed non-toxic fragment that reduces toxicity for this endpoint while keeping the overall scaffold as similar as possible.\\
- In step3\_reasoning, describe at a high level how the final molecule changes relative to the toxic molecule.\\
- Output the final non-toxic molecule as a single full SAFE string under the key ``answer''.\\
\\
Important:\\
- When modifying the toxic molecule to make it non-toxic, do not change other physicochemical or pharmacological properties; only reduce or remove the drug toxicity for this endpoint.\\
- Your output must be a SINGLE JSON object.\\
- Do not output any text outside the JSON.\\
- The fragment fields must be SAFE fragment strings (dot-separated if multiple).\\
\\
Output format: a single JSON object with the following keys:\\
- ``step1\_only\_toxic\_safe\_fragments'': string (dot-separated SAFE fragment(s))\\
- ``step1\_reasoning'': string\\
- ``step2\_only\_nontoxic\_safe\_fragments'': string (dot-separated SAFE fragment(s))\\
- ``step2\_reasoning'': string\\
- ``step3\_reasoning'': string\\
- ``answer'': string (the final nontoxic full SAFE string)\\
\\
Example:\\
{\ttfamily
\{"step1\_only\_toxic\_safe\_fragments":"frag1.frag2",\\
"step1\_reasoning":"...",\\
"step2\_only\_nontoxic\_safe\_fragments":"fragA.fragB",\\
"step2\_reasoning":"...",\\
"step3\_reasoning":"...",\\
"answer":"CCO.[*:1]"\}
}

\endgroup
\end{tcolorbox}
\captionsetup{skip=8pt, labelfont=bf}
\caption{Task 3 step-wise CoT question template for end-to-end non-toxic SAFE generation. Common context components are abbreviated as placeholders and described separately in the paper.}
\label{tab:task3_stepwise_cot_safe_question_prompt}
\end{table*}
\end{document}